\documentclass{article}

\usepackage[nonatbib, final]{neurips_data_2022}
\usepackage[numbers]{natbib}

\usepackage[utf8]{inputenc} 
\usepackage[T1]{fontenc}    
\usepackage{hyperref}       
\usepackage{url}            
\usepackage{booktabs}       
\usepackage{amsfonts}       
\usepackage{nicefrac}       
\usepackage{microtype}      
\usepackage{xcolor}         
\usepackage{graphicx}       
\usepackage{wrapfig}
\usepackage{svg}
\usepackage{amsmath}
\usepackage[inline]{enumitem}
\usepackage{multirow}

\newcommand{\NLE}{\texttt{NLE}}
\newcommand{\NLD}{\texttt{NLD}}
\newcommand{\NAO}{\texttt{NAO}}
\newcommand{\NLDobs}{\texttt{NLD-NAO}}
\newcommand{\NLDact}{\texttt{NLD-AA}}
\newcommand{\NLDmonk}{\texttt{NLD-AA-Monk}}

\newcommand{\AutoAscend}{\texttt{AutoAscend}}
\newcommand{\version}[3]{\texttt{v#1\kern-.05em.\kern-.05em#2\kern-.05em.\kern-.05em#3}}
\newcommand{\versionNH}[3]{\texttt{#1\kern-.05em.\kern-.05em#2\kern-.05em.\kern-.05em#3}}

\newcommand{\DatasetObject}{\texttt{TtyrecDataset}}
\newcommand{\ttyrec}{\texttt{ttyrec}}

\newcommand{\ttyrecthree}{\texttt{ttyrec3}}
\newcommand{\ttyrecbz}{\texttt{ttyrec\kern-.05em.\kern-.05embz2}}
\newcommand{\ttyrecthreebz}{\texttt{ttyrec3\kern-.05em.\kern-.05embz2}}
\newcommand{\xlogfile}{\texttt{xlogfile}}

\newcommand{\widecolpad}{\phantom{Kickstarting + BC(O))}& \phantom{ PADDIN WORD} & \phantom{  PADDIN WORD} & \phantom{(see column 1}}

\newcommand{\eg}{\textit{e.g.}}
\newcommand{\ie}{\textit{i.e.}}

\newcommand{\replychecklist}[1][]{\textcolor{blue}{[#1]}}
\title{Dungeons and Data: A Large-Scale NetHack Dataset }

\author{%
  Eric Hambro\thanks{Correspondence to \texttt{ehambro@fb.com}.} \\
  Meta AI\\
  \And
  Roberta Raileanu \\
  Meta AI \\
  \And
  Danielle Rothermel \\
  Meta AI \\
  \And
  Vegard Mella \\
  Meta AI \\
  \AND
  Tim Rockt\"aschel \\
  University College London\thanks{Work done while at Meta AI.}\\ 
  \And
  Heinrich K\"uttler \\
  Inflection~AI$^\dagger$
  \And
  Naila Murray \\
  Meta AI \\
}

\vspace{-3em}

\begin{document}

\maketitle

\vspace{-2em}

\begin{abstract}
Recent breakthroughs in the development of agents to solve challenging sequential decision making problems such as Go~\cite{alphago}, StarCraft~\cite{alphastar}, or DOTA~\cite{dota}, have relied on both simulated environments and large-scale datasets.
However, progress on this research has been hindered by the scarcity of open-sourced datasets and the prohibitive computational cost to work with them.
Here we present the NetHack Learning Dataset (\NLD{}), a large and highly-scalable dataset of trajectories from the popular game of NetHack, which is both extremely challenging for current methods and very fast to run~\cite{nle}. \NLD{} consists of three parts: 10 billion state transitions from 1.5 million human trajectories collected on the \NAO{} public NetHack server from 2009 to 2020; 3 billion state-action-score transitions from 100,000 trajectories collected from the symbolic bot winner of the NetHack Challenge 2021; and, accompanying code for users to record, load and stream any collection of such trajectories in a highly compressed form. 
We evaluate a wide range of existing algorithms including online and offline RL, as well as learning from demonstrations, showing that significant research advances are needed to fully leverage large-scale datasets for challenging sequential decision making tasks. 

\end{abstract}

\section{Introduction}

Recent progress on deep reinforcement learning (RL) methods has led to significant breakthroughs such as training autonomous agents to play Atari~\cite{atari}, Go~\cite{alphago}, StarCraft~\cite{alphastar}, and Dota~\cite{dota}, or to perform complex robotic tasks~\cite{Lee2020LearningQL, Rusu2017SimtoRealRL, Levine2018LearningHC, OpenAI2019SolvingRC, Heess2017EmergenceOL, Rajeswaran2018LearningCD}. In many of these cases, success relied on having access to large-scale datasets of human demonstrations~\cite{alphastar, dota, Rajeswaran2018LearningCD, Levine2018LearningHC}. Without access to such demonstrations, training RL agents to operate effectively in these environments remains challenging due to the hard exploration problem posed by their vast state and action spaces. In addition, having access to a simulator is key for training agents that can discover new strategies not exhibited by human demonstrations.
Therefore, training agents using a combination of offline data and online interaction has proven to be a highly successful approach for solving a variety of challenging sequential decision making tasks. 
However, this requires access to complex simulators and large-scale offline datasets, which tend to be computationally expensive. 

The NetHack Learning Environment (\NLE{}) was recently introduced as a testbed for RL, providing an environment which is both extremely challenging~\cite{hambro2022insights} and exceptionally fast to simulate~\cite{nle}.
\NLE{} is a stochastic, partially observed, and procedurally generated RL environment based on the popular game of NetHack. Due to its long episodes (\ie{} tens or hundreds of thousands of steps) and large state and action spaces, \NLE{} poses a uniquely hard exploration challenge for current RL methods. Thus, one of the most promising research avenues towards progress on NetHack is leveraging human or symbolic-bot demonstrations to bootstrap performance, which also proved successful for StarCraft~\cite{alphastar} and Dota~\cite{dota}.  

In this paper, we introduce the NetHack Learning Dataset (\NLD{}), an open and accessible dataset for large-scale offline RL and learning from demonstrations. \NLD{} consists of three parts: first, \NLDobs{}: a collection of state-only trajectories from 1.5 million human games of NetHack played on \texttt{nethack.alt.org (\NAO{})} servers between 2009 and 2020; second, \NLDact{}: a collection of state-action-score trajectories from 100,000 \NLE{} games played by the symbolic-bot winner of the 2021 NetHack Challenge~\cite{hambro2022insights}; third, \DatasetObject{}: a highly-scalable tool for efficient training on any NetHack and \NLE{}-generated trajectories and metadata. \NLD{}, in combination with \NLE{}, enables computationally-accessible research in multiple areas including imitation learning, offline RL, learning from sequences of only observations, as well as combining learning from offline data with learning from online interactions. In contrast with other large-scale datasets of demonstrations, \NLD{} is highly efficient in both memory and compute. \NLDobs{} can fit on a \$30 hard drive, after being compressed (by a factor of 160) from 38TB to 229GB. In addition, \NLDobs{} can be processed in under 15 hours, achieving a throughput of 288,000 frames per second with only 10 CPUs. \NLD{}'s low memory and computational requirements makes large-scale learning from demonstrations more accessible for academic and independent researchers. 

To summarize, the key characteristics of \NLD{} are that: it is a \textbf{scalable dataset of demonstrations} (\ie{} large and cheap) 
for a highly-complex sequential decision making challenge; it enables \textbf{research in multiple areas} such as imitation learning, offline RL, learning from observations of demonstrations, learning from both static data and environment interaction; and it has \textbf{many properties of real-world domains} such as partial observability, stochastic dynamics, sparse reward, long trajectories, rich environment, diverse behaviors, and a procedurally generated environment.


In this paper, we make the following core contributions:
\begin{enumerate*}[label=(\roman*)]
    \item we introduce \NLDobs{}, a large-scale dataset of almost 10 billion state transitions, from 1.5 million NetHack games played by humans;
    \item we also introduce \NLDact{}, a large-scale dataset of over 3 billion state-action-score transitions, from 100,000 games played by the symbolic-bot winner of the NeurIPS 2022 NetHack Competition;
    \item we open-source code for users to record, load,
    and stream any collection of NetHack trajectories in a highly compressed form; and
    \item we show that, while current state-of-the-art methods in offline RL and learning from demonstrations can effectively make use of the dataset, playing NetHack at human-level performance remains an open research challenge.  
\end{enumerate*} 

\section{Related Work}

\textbf{Offline RL Benchmarks.}
Recently, there has been a growing interest in developing better offline RL methods~\cite{Levine2020OfflineRL, Prudencio2022ASO, bcq, brac, brac+, cql, iql, atac, Agarwal2020AnOP, Siegel2020KeepDW} which aim to learn from datasets of trajectories. With it, a number of offline RL benchmarks have been released~\cite{Agarwal2019StrivingFS, rlunplugged, d4rl, rlds, Kurin2017TheAG, Ross2011ARO}. While these benchmarks focus specifically on offline RL, our datasets enable research on multiple areas including imitation learning, learning from observations only (\ie{} without access to actions or rewards), as well as learning from both offline and online interactions. In order to make progress on difficult RL tasks such as NetHack, we will likely need to learn from both human data and environment interaction, as was the case with other challenging games like StarCraft~\cite{alphastar} or Dota~\cite{dota}. In contrast, the tasks proposed in current offline RL benchmarks are much easier and can be solved by training either only online or only offline~\cite{d4rl, rlunplugged}. 
%
%
In addition, current offline RL benchmarks test agents on the exact same environment where the data was collected. As emphasized by~\cite{Toyer2020TheMB}, imitation learning algorithms drastically overfit to their environments, so it is essential to evaluate them on new scenarios in order to develop robust methods. In contrast, \NLE{} has long procedurally generated episodes which require both long-term planning and systematic generalization in order to succeed. 
This is shown in \cite{matthews2022hierarchical}, which investigates transfer learning between policies trained on different NLE-based environments.

For many real-world applications such as autonomous driving or robotic manipulation, learning from human data is essential due to safety concerns and time constraints~\cite{Qin2021NeoRLAN, d4rl, robonet, Cabi2019AFF, Li2010ACA, Strehl2010LearningFL, Thomas2017PredictiveOP, Henderson2008HybridRL, Pietquin2011SampleefficientBR, Jaques2019WayOB}. However, most offline RL benchmarks contain synthetic trajectories generated by either random exploration, pretrained RL agents, or simple hard-coded behaviors~\cite{d4rl}. In contrast, one of our datasets consists entirely of human replays, while the other one is generated by the winner of the NetHack Competition at NeurIPS 2022 which is a complex symbolic bot with built-in knowledge of the game. Human data (like the set contained in \NLDobs{}) is significantly more diverse and messy than synthetic data, as humans can vary widely in their expertise, optimize for different objectives (such as fun or discovery), have access to external information (such as the NetHack Wiki~\cite{nhwiki}), and even have different observation or action spaces than RL agents. Hence, learning directly from human data is essential for making progress on real-world problems in RL. 

\textbf{Large-Scale Human Datasets.}
A number of large-scale datasets of human replays have been released for StarCraft~\cite{starcraft2}, Dota~\cite{dota}, and MineRL~\cite{minerl}. However, training models on these datasets requires massive computational resources, which makes it unfeasible for academic or independent researchers. In contrast, \NLD{} strikes a better balance between scale (\ie{} a large number of diverse human demonstrations on a complex task) and efficiency (\ie{} cheap to use and fast to run).

For many real-world applications such as robotic manipulation, we only have access to the demonstrator's observations and not their actions~\cite{Mandlekar2021WhatMI, Sharma2018MultipleIM, Qin2021NeoRLAN, robonet, Stadie2017ThirdPersonIL}. Research on this setting has been slower~\cite{ilpo, bco, cohen2021imitation}, in part due to the lack of efficient large-scale datasets. While there are some datasets containing only observations, they are either much smaller than \NLD~\cite{Memmesheimer2019SimitateAH, Toyer2020TheMB, Sharma2018MultipleIM}, too computationally expensive~\cite{starcraft2, dota}, or lack a simulator which prevents learning from online interactions~\cite{Geiger2013VisionMR, Mahler2019LearningAR, robonet}. 




\section{Background: The NetHack Learning Environment}

The NetHack Learning Environment (\NLE{}) is a \textit{gym} environment~\cite{brockman2016openai} based on the popular “dungeon-crawler” game, NetHack~\cite{NetHackOrg}. Despite the visual simplicity,
NetHack is widely considered one of the hardest video games in history since it can take years for humans to win the game~\cite{telegraph}. Players need to explore the dungeon, manage their resources, as well as learn about the many entities and their dynamics (often by relying on external knowledge sources like the NetHack Wiki~\cite{nhwiki}). NetHack has a clearly defined goal, namely descend the dungeon, retrieve an amulet, and ascend back to win the game. At the beginning of each game, players are randomly assigned a given multidimensional character defined by role, race, alignment, and gender (which have varying properties and challenges), so they need to master all characters in order to win consistently. Thus, \NLE{} offers a unique set of properties which make it well-positioned to advance research on RL and learning from demonstrations: it is a highly complex environment, containing hundreds of entities with different dynamics; it is procedurally generated, allowing researchers to test generalization; it is partially observed, highly stochastic, and has very long episodes (\ie{} one or two orders of magnitude longer that Starcraft II~\cite{starcraft2}). 

Following its release, several works have built on \NLE{} to leverage its complexity in different ways. MiniHack~\cite{samvelyan2021minihack} allows researchers to design their own environments to test particular capabilities of RL agents, by leveraging the full set of entities and dynamics from NetHack. The NetHack Challenge~\cite{hambro2022insights} was a competition at NeurIPS 2021, which sought to incentivise a showdown between symbolic and deep RL methods on \NLE{}. Symbolic bots decisively outperformed deep RL methods, with the best performing symbolic bots surpassing state-of-the-art deep RL methods by a factor of 5.

\section{The NetHack Learning Dataset}

The NetHack Learning Dataset (\NLD{}) contains three components: 
\begin{enumerate}
    \item \NLDobs{} --- a directory of \ttyrecbz{} files containing almost 10 billion state trajectories and metadata from 1,500,000 human games of NetHack played on \texttt{nethack.alt.org}.
    \item \NLDact{} --- a directory of \ttyrecthreebz{}  files containing over 3 billion state-action-score trajectories and metadata from 100,000 games collected from the winning bot of the NetHack Challenge~\cite{hambro2022insights}. 
    \item \DatasetObject{} --- a Python class that can scalably load directories of \ttyrecbz{} / \ttyrecthreebz{} files and their metadata into \texttt{numpy} arrays.
\end{enumerate}
We are also releasing a new version of the NetHack environment, \NLE{} \version{0}{9}{0}, which contains new features and ensures compatibility with \NLD{} (see Appendix~\ref{sec:nle_version}). 

\textbf{File Format.}
The \ttyrec{}\footnote{\url{https://nethackwiki.com/wiki/Ttyrec}} file format stores sequences of terminal instructions (equivalent to observations in RL), along with the times at which to play them.
In \NLE{} \version{0}{9}{0}, we adapt this format to also store keypress inputs to the terminal (equivalent to actions in RL), and in-game scores over time (equivalent to rewards in RL), allowing a reconstruction of state-action-score trajectories. This adapted format is known as \ttyrecthree{}.  
The \ttyrecbz{} and \ttyrecthreebz{} formats, compressed versions of \ttyrec{} and \ttyrecthree{}, are the primary data formats used in \NLD{}. 
Using \DatasetObject{}, these compressed files can be written and read on-the-fly, resulting in data compression ratios of more than 138.
The files can be decompressed into the state trajectory on a terminal, by using a terminal emulator and querying its screen. For more details see Appendix \ref{sec:ttyrec}.

Throughout the paper, we refer to a player's input at a given time as either state or observation. However, note that NetHack is partially observed, so the player doesn't have access to the full state of the game. We also sometimes use the terms score and reward interchangeably, since the increment in in-game score is a natural choice for the reward used to train RL agents on NetHack. Similarly, a human's keypress corresponds to an agent's action in the game.

\textbf{Metadata.}
NetHack has an optional built-in feature for the logging of game metadata, used for the maintenance of all-time high-score lists. At the end of a game, 26 fields of data are logged to a common \xlogfile{}\footnote{\url{https://nethackwiki.com/wiki/Xlogfile}} for posterity. These fields include the character's \textit{name}, \textit{race}, \textit{role}, \textit{score}, \textit{cause of death}, \textit{maximum dungeon depth}, and more. See Appendix~\ref{sec:metadata} for more details on these fields. With \NLE{} \version{0}{9}{0}, an \xlogfile{} is generated for all \NLE{} recordings. These files are used to populate all metadata contained in \NLD{}. 

\textbf{State-Action-Score Transitions.}
As mentioned, \NLDact{} contains sequences of state-action-score transitions from symbolic-bot plays, while \NLDobs{} contains sequences of state transitions from human plays. These transitions are efficiently processed using the \DatasetObject{}. The states consist of: \texttt{tty\_chars} (the characters at each point on the screen), \texttt{tty\_colors} (the colors at each point on the screen), \texttt{tty\_cursor} (the location of the cursor on the screen), \texttt{timestamps} (when the state was recorded), and \texttt{gameids} (an identifier for the game in question).  Additionally, \texttt{keypresses} and  \texttt{score} observations are available for \ttyrecthree{} files, as in the \NLDact{} dataset. The states, keypresses, and scores from \NLD{} map directly to an agent's observations, actions, and rewards in \NLE{}. More information about these transitions can be found in Appendix~\ref{sec:observation}. 

\textbf{API.}
The \DatasetObject{} follows the API of an \texttt{IterableDataset} defined in PyTorch. This allows for the batched streaming of \ttyrecbz{} / \ttyrecthreebz{} files directly into fixed NumPy arrays of a chosen shape.  Episodes are read sequentially in chunks defined by the unroll length, batched with a different game per batch index. The order of these games can be shuffled, predetermined or even looped to provide a never-ending dataset. 
This class allows users to load any state-action-score trajectory recorded from  \NLE{} \version{0}{9}{0} onwards.

The \DatasetObject{} wraps a small sqlite3 database where it stores metadata and the paths to files.  This design allows for the simple querying of metadata for any game, along with the dynamic subselection of games streamed from the \DatasetObject{} itself. For example, in Figure \ref{fig:scoreboxplots}, we generate sub-datasets from \NLDobs{} and \NLDact{}, selecting trajectories where the player has completed the game (`Ascended') or played a `Human Monk' character, respectively. Appendix \ref{sec:api} shows how to load only a subset of trajectories, for example where the player has ascended, or a certain role has been used. 

\textbf{Scalability.}
The \DatasetObject{} is designed to make our large-scale \NLD{} datasets accessible even when the computational resources are limited. 
%
To that end, several optimizations are made to improve the memory efficiency and throughput of the data.  
Most notably, \DatasetObject{} streams recordings directly from their compressed \ttyrecbz{} files. This format compresses the 30TB of frame data included in \NLDobs{} down to 229GB, which can fit on a \$30 SSD\footnote{https://www.amazon.com/HP-240GB-Internal-Solid-State/dp/B09KFHTYWH}. This decompression requires on-the-fly unzipping and terminal emulation. The \DatasetObject{} performs these in GIL-released C/C++. This process is fast and trivially parallelizable with Python Threadpool, resulting in throughputs of over 1.7 GB/s on 80 CPUs. This performance allows the processing of almost 10 billion frames of \NLDobs{} in 15 hours, with 10 CPUs. See Table~\ref{tab:overview} for a quantitative description of our two datasets.

\begin{table}[]
    \centering
    \caption{\NLDact{} and \NLDobs{} in numbers.}
    \begin{tabular}{l l l}
     \toprule
    & \NLDact{} & \NLDobs{}  \\
     \midrule
        Episodes & 109,545 &  1,511,228 \\
        Transitions & 3,481,605,009 &  9,858,127,896 \\
        Policies (Players)  & 1 & 48,454 \\
        Policies Type  & symbolic bot & human \\
        Transition     & (state, action, score) & state \\
        \midrule
        Disk Size (Compressed) & 96.7 GB & 229 GB\\
        Data Size (Uncompressed) & 13.4 TB & 38.0 TB \\
        Compression Ratio & 138 & 166 \\  
 
        \midrule
        Mean Epsiode Score & 10,105 & 127,218\\
        Median Episode Score & 5,422 & 836 \\
        Median Episode Transitions & 28,181 &  1,724.0 \\
        Median Episode Game Turns & 20,414 & 3,766 \\
        \midrule       
        Epoch Time (10 CPUs)    &  4h 49m  & 14h 37m \\ 
        \bottomrule
     \end{tabular}
    \label{tab:overview}
     \vspace{-1.5em}
\end{table}

\section{Dataset Analysis}
\label{sec:analysis}
In this section we perform an in-depth analysis of the characteristics of \NLDact{} and \NLDobs{}.

\subsection{\NLDact{}}

To our knowledge, \AutoAscend{}\footnote{\url{https://github.com/maciej-sypetkowski/autoascend}} is currently the best open-sourced artificial agent for playing NetHack \versionNH{3}{6}{6}, having achieved first place in the 2021 NeurIPS NetHack Challenge by a considerable margin~\cite{hambro2022insights}. 
\texttt{AutoAscend} is a symbolic bot, forgoing any neural network and instead relying on a priority list of hard-coded subroutines. 
These subroutines are complex, context dependant, and make significant use of NetHack domain knowledge and all NetHack actions. For instances, the bot keeps track of multiple properties for encountered entities and can even solve challenging puzzles such as Sokoban. A full description of \AutoAscend{}'s algorithm and behavior can be found in the NetHack Challenge Report \cite{hambro2022insights}.

\NLDact{} was generated by running \AutoAscend{} on the \texttt{NetHackChallenge-v0}~\cite{hambro2022insights} task in \NLE{} \version{0}{9}{0},  utilising its built-in recording feature to generate \ttyrecthreebz{} files.  
It consists of over 3 billion state-action-score transitions, drawn from 100,000 episodes generated by \AutoAscend{}. 
Of the \NLE{} tasks, \texttt{NetHackChallenge-v0} most closely reproduces the full game of NetHack \versionNH{3}{6}{6}, and was introduced in \NLE{} \version{0}{7}{0} to grant NetHack Challenge competitors access to the widest possible action space, and force an automated randomisation of the starting character (by race, role, alignment, and gender). By virtue of using \ttyrecthreebz{} files, in-game scores and actions (in the form of keypresses) are stored along with the states, and metadata about the episodes.

\begin{figure}[t]
  \centering
  \includegraphics[width=0.8\linewidth]{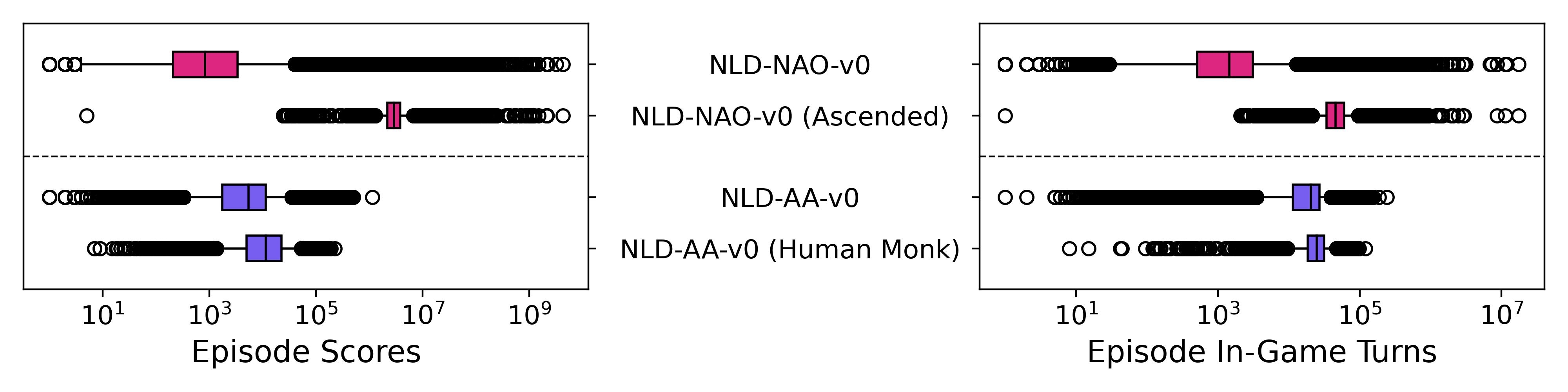}
  \caption{Box plots of episode scores (left) and episode turns (right) for \NLDobs{}} and \NLDact{} as well as for their corresponding subsets that contain only winning games (`Ascended') or games played with a `Human Monk' character, respectively. These plots show the distribution in ability in \NLDobs{} is heavy-tailed, along with the typical scores and game length needed to win, well above the median. By comparison,  \NLDact{} games are competent but not expert, even for easy characters like `Human Monk'. 
  \label{fig:scoreboxplots}
  \vspace{-1em}
\end{figure}

\textbf{Dataset Skill.}
The \AutoAscend{} trajectories demonstrate a strong and reliable NetHack player, far exceeding all the deep learning based approaches, but still falling short of an expert human player. NetHack broadly defines a character with less than 2000 score as a `Beginner'\footnote{\url{https://nethackwiki.com/wiki/Beginner}}. \AutoAscend{} comfortably surpasses the `Beginner' classification in more than 75\% of games for all roles, and in 95\% of games for easier roles like Monk (see Figure~\ref{fig:scoreboxplots}). Given the high variance nature of NetHack games, and the challenge of playing with the more difficult roles, this is an impressive feat. 

Compared to the human players in \NLDobs{}, \AutoAscend{}'s policy finds itself just within the top 15\% of all players when ranked by mean score. When ranked by median score\footnote{Median score was the primary metric in the NetHack Challenge} it comes within the top 7\%.  However, these metrics are somewhat distorted by the long tail of dilettante players who played only a few games. If instead we define a `competent' human player as one to have \textit{ever} advanced beyond the Beginner classification, then \AutoAscend{} ranks in the top 33\% of players by mean score, and top 15\% by median.

The competence of \AutoAscend{} contrasts both with the poor performance of deep RL bots, and with the exceptional performance required to beat the game.  As the winning symbolic bot of the NetHack Challenge, \AutoAscend{} beat the winning deep learning bot by almost a factor of 3 in median score, and close to a factor of 5 in mean score.  This performance is far outside what deep RL agents can currently achieve, and in some domains \NLDact{} may be considered an ``expert" dataset.  However, when compared to the distribution of winning games (\ie{} those that have `ascended'), both mean and median score are orders of magnitude short.

\textbf{Dataset Coverage.} 
Thanks to its scale, \NLDact{} provides solid coverage of the early-to-mid game of NetHack. \NLDact{} also provides good coverage of all possible actions and roles, with episodes that last for tens of thousands of transitions. Nevertheless, the dataset contains many complex and diverse behaviors. For the full breakdown of the symbolic bot's behaviors found in \NLDact{} see Appendix \ref{sec:metadata}. 


\textbf{Datasest Noise.} 
Since all trajectories were generated using \NLE{}, these trajectories have no terminal rendering noise. \NLE{} uses a VT100 emulator to record \ttyrecthreebz{} files, and \NLD{} uses the same emulator to load them.



\subsection{\NLDobs{}}

The NetHack community has a long history of using public servers, wherein players can play together, stream and rewatch games of NetHack.\footnote{\url{https://nethackwiki.com/wiki/Public\_server}}. 
The \url{nethack.alt.org} server, known as \NAO{}\footnote{\url{https://nethackwiki.com/wiki/Nethack.alt.org}}, is one of the longest running of such servers, having started in 2001, and hosts recordings dating back to 2009. 
The server attempts to record all games in \ttyrecbz{} format, and hosts them publicly on an S3 bucket\footnote{\url{s3://altorg/ttyrec/}}. 
Unlike \NLE{}, \NAO{} offers players the option of saving a game and resuming play later, allowing for games that span multiple sessions, or last potentially days, weeks, months or years.

\NLDobs{} was collected by downloading and filtering a dump of \ttyrecbz{} recordings and metadata from NAO.
It consists of almost 10 billion state transitions, drawn from approximately 1.5 million games played by just under 50,000 humans of mixed ability.
These games were played between 2009 and 2020, and span NetHack versions \version{3}{4}{0} to \versionNH{3}{6}{6}.
Since \NAO{} provides no link between episode metadata and available \ttyrec{} recordings, we developed an algorithm to assign and correctly order \ttyrec{} recordings to their corresponding episode metadata, and to filter out trivial or empty games. The full procedure is outlined in Appendix~\ref{sec:appendix-nao}.

\textbf{Dataset Skill.}
The \NAO{} trajectories were generated from humans of very different skill levels, ranging from novice to expert, and cover a vast distribution of outcomes.  
The distributions of episode scores, episode turns, and average player scores are highly Zipfian, with long correlated tails spanning several orders of magnitude.  
The correlation of these distributions greatly improves the quality of the dataset. While the median episode score does not progress beyond Beginner status, such short episodes contribute relatively few transitions to the overall dataset.
The bulk of the transitions comes from medium-to-high scoring games that last a long time. 
The \NLDobs{} dataset therefore combines two desirable properties: it contains a large number of diverse behaviors with varying levels of performance, as well as many good trajectories from expert demonstrators. 

\textbf{Dataset Coverage.} 
Thanks to its size and expert policies, \NLDobs{} provides an unparalleled coverage of the game of NetHack. The dataset contains many examples of advanced game sections not yet achievable by symbolic or RL bots, such as descending the dungeon and retrieving the amulet. Most importantly, it includes 22,000 examples of winning the game (``Ascension''), and includes several examples of doing so with special conducts such as ``Nudist'' (not wearing any armour). The full distribution of such achievements and conducts can be found in Appendix~\ref{sec:metadata}.


\textbf{Dataset Noise.} 
Unlike \NLE{}, \NAO{} does not enforce any fixed terminal type on the player, nor any fixed screen dimensions or configuration. Variations in all these can lead to rendering noise that may include odd symbols or cropped layouts. These types of artefacts are common in real-world applications of learning from demonstrations, which makes \NLD{} a suitable dataset for advancing research in this area. More details about these artefacts, as well as other types of noise (\eg{} matching of recordings to players), see Appendix~\ref{sec:appendix-nao}.


\section{Experiments}

\subsection{Methods} 


We demonstrate the utility and challenge of \NLD{} by evaluating algorithms from a range of fields including  online RL, offline RL, imitation learning, and learning from demonstrations. 

All methods are implemented using the open-source asynchronous RL library \texttt{moolib}~\cite{moolib2022}, allowing for continual evaluation of agents in order to generate training curves (even in the offline RL setting). In addition, all our methods use the same core model architecture, which is based on a popular open-source Sample Factory baseline\footnote{\url{https://github.com/Miffyli/nle-sample-factory-baseline}}, adapted to only use \texttt{tty\_*} observations but still remain fast to run. Unless specified otherwise, all experiments were evaluated with 1024 episodes from \NLE{}'s \textit{NetHackChallenge-v0} task after training for 1.5B frames. The mean and standard deviation of the episode return was computed over 10 training seeds. Full details about the algorithms and their hyperparameters can be found in Appendix~\ref{sec:appendix-experiments}. All our implementations will be open-sourced to enable fast research progress.

\textbf{Online RL.} For online RL baselines that do not make use of the \NLD{} datasets, we use Asynchronous Proximal Policy Optimization (APPO)~\cite{schulman2017proximal, petrenko2020sample} and Deep Q-Networks (DQN-Online). The former is a strong on-policy policy-gradient-based method for \NLE{}, and the latter is a popular off-policy value-based method in the wider literature.

\textbf{Offline RL.} Since \NLDact{} contains state-action-reward transitions, we can train a number of offline RL baselines on it. More specifically, we use a Deep Q-Network baseline trained on the dataset (DQN-Offline), as well as two state-of-the-art offline RL methods, Conservative Q-Learning (CQL)~\cite{cql} and Implicit Q-Learning (IQL)~\cite{iql}. Note that these methods cannot be trained on \NLDobs{} since it doesn't contain actions or rewards.  

\textbf{Learning from Demonstrations.} We also evaluate a number of baselines that do not make use of rewards, but instead use state-action or state-only transitions. For \NLDact{}, we use Behavioural Cloning (BC), a popular imitation learning method, which trains a policy to match the corresponding actions in the dataset using a supervised learning objective. For \NLDobs{}, we use Behavioral Cloning from Observations (BCO)~\cite{bco} which is a popular baseline for learning from state-only demonstrations. This method performs BC on a learned inverse dynamics model which was trained to predict the action taken given two consecutive states. We then investigate the use of BC and BCO in an online setting by augmenting our APPO baseline in two ways. First, we add the BC or BCO loss between the agent's policy and the expert's action for observations in the dataset resulting in APPO + BC and APPO + BCO respectively. Second, we add a KL-divergence loss between the agent's policy and a pretrained BC(O) model resulting in Kickstarting BC(O), similar to~\cite{kickstarting}. These methods make use of state and action data but no rewards. 


For \NLDact{}, we investigate the performance of the above algorithms across two partitions: one with a single character (‘Human Monk’), and one with all the characters (\ie{} the full dataset). Each of these is evaluated on their corresponding environments. Full details about our experimental setup are available in Appendix~\ref{sec:appendix-experiments}.


\begin{table}[]
    \caption{Average episode return across 10 seeds for a number of popular baselines using \NLD{} and / or \NLE{}. First, we evaluate a range of offline RL methods such as DQN-Offline, CQL, IQL and BC trained on \NLDact{} and \NLDmonk{}, as well as BCO trained on \NLDobs{}. Second, we evaluate methods that combine offline and online learning such as Kickstarting + BC and APPO + BC trained on \NLDact{} and \NLDmonk{}, as well as Kickstarting + BCO and APPO + BCO trained on \NLDobs{}. Finally, we evaluate a few methods that only have access to online interactions from \NLE{} (no external datasets) such as a Random Policy, DQN-Online, and APPO. All evaluations are conducted with \NLE{}'s \textit{NetHackChallenge-v0} task, using \texttt{Random} (\texttt{@}) characters for \NLDact{} and \NLDobs{}, and a \texttt{Human Monk} character for \NLDmonk{}. We compare all these to the average episode return in the corresponding datasets. Methods that use both offline and online data outperform all others by a wide margin, but are still significantly worse than humans (by almost two orders of magnitude) and even the symbolic bot (by more than a factor of three). The top online algorithm, APPO, outperforms the top offline approach, BC. These results indicate that \NLD{} poses a substantial challenge to state-of-the-art methods.}
     \centering
     \hskip-2.0cm
     \begin{tabular}{r@{}c}
                    & $\begin{tabular}{@{}lccc@{}}
                      \toprule
                                                   & \NLDact{}                  & \NLDmonk{}          &  \NLDobs{}      \\
                       Character                   & \texttt{Random}            & \texttt{Human Monk} &  \texttt{Random} \\
                       \midrule  
                       \widecolpad{} \\[-\normalbaselineskip]
                       \end{tabular} \kern-\nulldelimiterspace$  \\
       offline only & $\left\{
                       \begin{tabular}{@{}lccc@{}}
                        DQN-Offline       &       0.0  ± 0.0            & 0.0 ± 0.0        &  -    \\
                        CQL               &        33 ± 70            & 56 ± 89        &  -   \\
                        IQL               &        160 ± 70             & 201 ± 84        &  - \\
                        \textbf{BC(O)}   & \textbf{554 ± 142}  & \textbf{1058 ± 503}  &   \textbf{5.4 ± 2.7} \\
                       \widecolpad{} \\[-\normalbaselineskip]%
                       \end{tabular}\right.\kern-\nulldelimiterspace$ \\
                    & $\begin{tabular}{@{}lccc@{}}
                       \midrule 
                       \widecolpad{}\\[-\normalbaselineskip]
                       \end{tabular} \kern-\nulldelimiterspace$ \\
   offline + online & $\left\{
                       \begin{tabular}{@{}lccc@{}}
                        Kickstarting + BC(O)     &         962 ± 158   &         2090 ± 388  &          245 ± 125 \\
                        \textbf{APPO + BC(O)}  & \textbf{1282 ± 275}   & \textbf{2809 ± 324} & \textbf{884 ± 117}\\

                        \widecolpad{}\\[-\normalbaselineskip]%
                        \end{tabular}\right.\kern-\nulldelimiterspace$ \\
                    & $\begin{tabular}{@{}lccc@{}}
                       \midrule 
                       \widecolpad{}\\[-\normalbaselineskip]
                       \end{tabular} \kern-\nulldelimiterspace$ \\
        online only & $\left\{
                       \begin{tabular}{@{}lccc@{}}
                       Random Policy        &    0.0 ± 0.0                & 0.1 ± 0.3         &  (see col 1)  \\
                       DQN-Online   &             89 ± 46               &  130 ± 150           & (see col 1)\\
                        \textbf{APPO} &  \textbf{868 ± 200}     &  \textbf{2046 ± 360} & \textbf{(see col 1)} \\
                       \widecolpad{}\\[-\normalbaselineskip]%
                       \end{tabular}\right.\kern-\nulldelimiterspace$ \\
                    & $\begin{tabular}{@{}lccc@{}}
                       \midrule
                       \textbf{Dataset Average}  & \textbf{10105}           & \textbf{17274}       & \textbf{127356} \\
                       \bottomrule 
                       \widecolpad{}\\[-\normalbaselineskip]
                       \end{tabular} \kern-\nulldelimiterspace$ \\
     \end{tabular}
    \label{tab:results}
      \vspace{-2em}
\end{table}

\subsection{Results and Discussion}
\label{resultsanddiscussion}
\begin{figure}

    \centering
    \includegraphics[width=\linewidth]{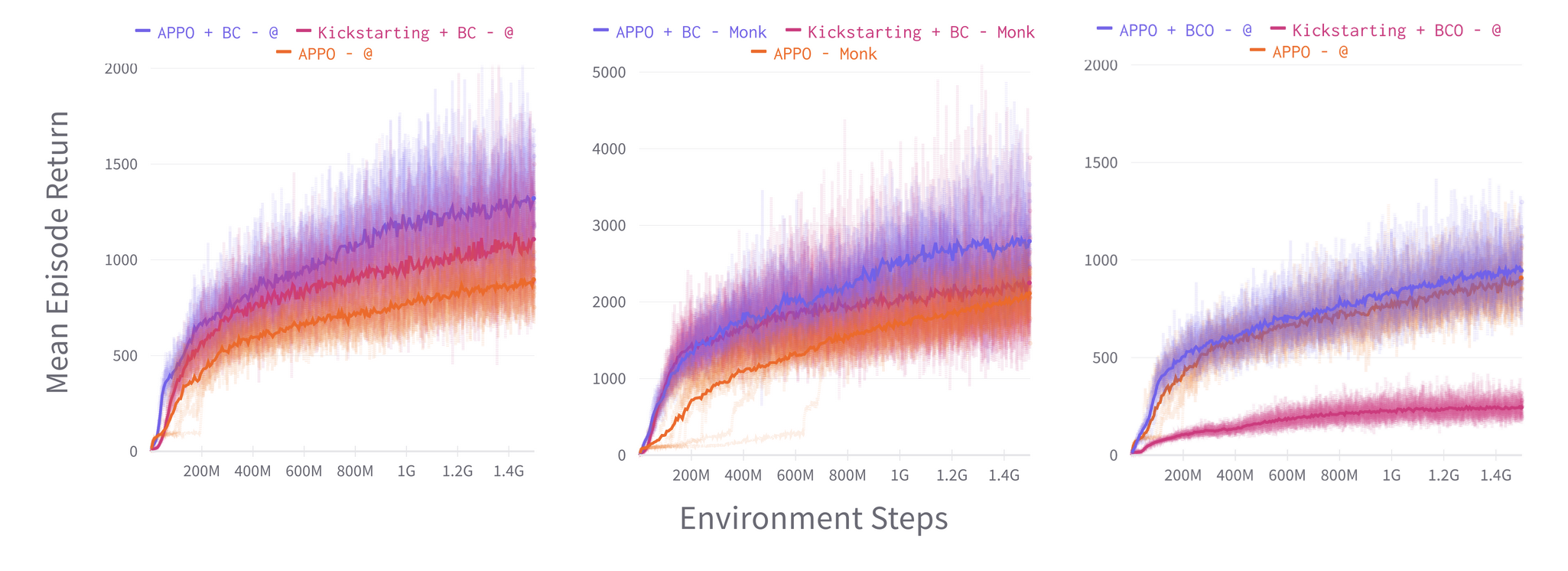}
    \vspace{-2.5em}
    \caption{Mean episode return (with samples plotted) over 10 seeds during training for APPO, APPO + BC(O), and Kickstarting + BC(O) using \NLDact{} (left), \NLDmonk{} (center), and \NLDobs{} (right). \NLDact{} and \NLDobs{} both are evaluated with \texttt{Random} character (\texttt{@}), while \NLDmonk{} is evaluated with only the \texttt{Human Monk} character. Some sample curves associated with the pure online APPO methods exhibit a behaviour wherein performance appears to plateau, before bursting rapidly to a new performance level. This is a common pattern in \NLE{} and corresponds to an agent learning a new behaviour that greatly unlocks yet more of the game, such as praying when starving or kicking down locked doors. Since \NLE{} is highly procedurally generated with a vast action and state space, it is hard to predict when an agent will encounter sufficient opportunities to ``take off'' in this way. Such ``taking off'' phenomena are less visible when learning from \NLD{}, as useful behaviours are present in the data throughout.}
    \label{fig:training_curves}
    \vspace{-1em}
\end{figure}

As Table~\ref{tab:results} and Figure~\ref{fig:training_curves} show, \NLDact{} poses a difficult challenge to modern offline RL algorithms. While CQL is able to use the offline trajectories to outperform both a random policy, IQL, as well as online and offline variants of DQN, it falls short of capturing the average performance of trajectories in the dataset by over an order of magnitude. BC performs better due to the higher quality of the supervised learning signal, which is typically denser and less noisy than the RL one, but it too falls well short of the data generating policy. Prior work has shown that offline RL methods struggle to perform well in stochastic environments~\cite{paster2022you}, and to generalize outside the distribution of state-action pairs in the dataset~\cite{gulcehre2020addressing}. These challenges are particularly intensified with \NLD{}, which is highly stochastic, partially observed, procedurally generated, and has a very large state and action space. 

%

Methods that combine offline and online training such as Kickstarting + BC(O) or APPO + BC(O) outperform offline-only methods and outperform or match online-only methods, while also speeding up training.
Despite being our best method, APPO + BC is still significantly worse than the average performance in the dataset, further emphasizing the difficulty of \NLD{}.



Finally, our results demonstrate the utility of \NLDobs{} while highlighting the challenge of learning from sequences of only observations. Our inverse model is trained with trajectories from a random policy, and therefore experiences an extremely small fraction of the entire state-action space. Even so, BCO on \NLDobs{} labeled with these actions is able to yield performance that comfortably exceeds that of a random policy. While APPO + BCO performs similarly to APPO, training appears to be much more stable. 
Thus, using more sophisticated policies for training the inverse model (\eg{} a mix between a pretrained RL policy and a random policy or an exploration method that maximizes the diversity of visited states and actions) could further improve performance.  

\section{Conclusion}
In this paper, we introduce \NLD{}, a large-scale dataset of demonstrations from NetHack, a complex, stochastic, partially observed, procedurally generated video game. \NLD{} enables research in multiple areas such as imitation learning, offline RL, learning from sequences of observations, and learning from both offline data and online interactions. \NLD{} allows researchers to learn from very large datasets of demonstrations without requiring extensive memory or compute resources. We train a number of state-of-the-art methods in online RL (both on-policy and off-policy), offline RL, imitation learning, as well as learning from observation-only sequences. Our results indicate that significant research advances are needed to fully leverage large-scale datasets in order to solve challenging sequential decision making problems. 

In many real-world applications such as robotics, it is common to have access to datasets of observations, without corresponding actions or rewards. Thus, one important direction for future research is to learn from trajectories containing only observations. \NLD{} enables this type of research in a safe, cheap, fast, and reproducible environment, while still exhibiting many properties of real-world domains such as partial observability, high stochasticity, and long episodes. Due to NetHack being procedurally generated, in order to consistently win the game, an agent needs to learn a robust policy that generalizes to new scenarios. Our strongest baseline learns from both static data and environment interaction, so a promising avenue for future work is to develop more effective methods that combine offline and online learning. Existing algorithms for offline RL and learning from demonstrations struggle to match the average performance of trajectories in our datasets, suggesting that \NLD{} is a good benchmark to probe the limits of these methods and inspire the development of approaches that are effective in more challenging and realistic domains.  

\section{Acknowledgements}

We would like to thank Drew Streib and the NAO maintainers for their work in the NetHack community and their help throughout this project. We would also like to thank Micha\l{} Sypetkowski and Maciej Sypetkowski for their work on the AutoAscend bot.

\bibliographystyle{plain}
\bibliography{ref}

\begin{thebibliography}{10}

\bibitem{Agarwal2019StrivingFS}
Rishabh Agarwal, Dale Schuurmans, and Mohammad Norouzi.
\newblock Striving for simplicity in off-policy deep reinforcement learning.
\newblock {\em ArXiv}, abs/1907.04543, 2019.

\bibitem{Agarwal2020AnOP}
Rishabh Agarwal, Dale Schuurmans, and Mohammad Norouzi.
\newblock An optimistic perspective on offline reinforcement learning.
\newblock In {\em ICML}, 2020.

\bibitem{dota}
Christopher Berner, Greg Brockman, Brooke Chan, Vicki Cheung, Przemyslaw
  Debiak, Christy Dennison, David Farhi, Quirin Fischer, Shariq Hashme,
  Christopher Hesse, Rafal J{\'o}zefowicz, Scott Gray, Catherine Olsson,
  Jakub~W. Pachocki, Michael Petrov, Henrique~Pond{\'e} de~Oliveira~Pinto,
  Jonathan Raiman, Tim Salimans, Jeremy Schlatter, Jonas Schneider, Szymon
  Sidor, Ilya Sutskever, Jie Tang, Filip Wolski, and Susan Zhang.
\newblock Dota 2 with large scale deep reinforcement learning.
\newblock {\em ArXiv}, abs/1912.06680, 2019.

\bibitem{brockman2016openai}
Greg Brockman, Vicki Cheung, Ludwig Pettersson, Jonas Schneider, John Schulman,
  Jie Tang, and Wojciech Zaremba.
\newblock Openai gym.
\newblock {\em arXiv preprint arXiv:1606.01540}, 2016.

\bibitem{Cabi2019AFF}
Serkan Cabi, Sergio~Gomez Colmenarejo, Alexander Novikov, Ksenia Konyushkova,
  Scott~E. Reed, Rae Jeong, Konrad Zolna, Yusuf Aytar, David Budden, Mel
  Vecer{\'i}k, Oleg~O. Sushkov, David Barker, Jonathan Scholz, Misha Denil,
  Nando de~Freitas, and Ziyun Wang.
\newblock A framework for data-driven robotics.
\newblock {\em ArXiv}, abs/1909.12200, 2019.

\bibitem{atac}
Ching-An Cheng, Tengyang Xie, Nan Jiang, and Alekh Agarwal.
\newblock Adversarially trained actor critic for offline reinforcement
  learning.
\newblock {\em ArXiv}, abs/2202.02446, 2022.

\bibitem{cohen2021imitation}
Samuel Cohen, Brandon Amos, Marc~Peter Deisenroth, Mikael Henaff, Eugene
  Vinitsky, and Denis Yarats.
\newblock Imitation learning from pixel observations for continuous control.
\newblock In {\em Deep RL Workshop NeurIPS 2021}, 2021.

\bibitem{robonet}
Sudeep Dasari, Frederik Ebert, Stephen Tian, Suraj Nair, Bernadette Bucher,
  Karl Schmeckpeper, Siddharth Singh, Sergey Levine, and Chelsea Finn.
\newblock Robonet: Large-scale multi-robot learning.
\newblock {\em ArXiv}, abs/1910.11215, 2019.

\bibitem{ilpo}
Ashley~D. Edwards, Himanshu Sahni, Yannick Schroecker, and Charles~Lee Isbell.
\newblock Imitating latent policies from observation.
\newblock {\em ArXiv}, abs/1805.07914, 2019.

\bibitem{d4rl}
Justin Fu, Aviral Kumar, Ofir Nachum, G.~Tucker, and Sergey Levine.
\newblock D4rl: Datasets for deep data-driven reinforcement learning.
\newblock {\em ArXiv}, abs/2004.07219, 2020.

\bibitem{bcq}
Scott Fujimoto, David Meger, and Doina Precup.
\newblock Off-policy deep reinforcement learning without exploration.
\newblock In {\em ICML}, 2019.

\bibitem{Geiger2013VisionMR}
Andreas Geiger, Philip Lenz, Christoph Stiller, and Raquel Urtasun.
\newblock Vision meets robotics: The kitti dataset.
\newblock {\em The International Journal of Robotics Research}, 32:1231 --
  1237, 2013.

\bibitem{gulcehre2020addressing}
Caglar Gulcehre, Sergio~G{\'o}mez Colmenarejo, Jakub Sygnowski, Thomas Paine,
  Konrad Zolna, Yutian Chen, Matthew Hoffman, Razvan Pascanu, Nando de~Freitas,
  et~al.
\newblock Addressing extrapolation error in deep offline reinforcement
  learning.
\newblock 2020.

\bibitem{minerl}
William~H. Guss, Brandon Houghton, Nicholay Topin, Phillip Wang, Cayden Codel,
  Manuela~M. Veloso, and Ruslan Salakhutdinov.
\newblock Minerl: A large-scale dataset of minecraft demonstrations.
\newblock {\em ArXiv}, abs/1907.13440, 2019.

\bibitem{hambro2022insights}
Eric Hambro, Sharada Mohanty, Dmitrii Babaev, Minwoo Byeon, Dipam Chakraborty,
  Edward Grefenstette, Minqi Jiang, Daejin Jo, Anssi Kanervisto, Jongmin Kim,
  et~al.
\newblock Insights from the neurips 2021 nethack challenge.
\newblock {\em arXiv preprint arXiv:2203.11889}, 2022.

\bibitem{Heess2017EmergenceOL}
Nicolas Manfred~Otto Heess, TB~Dhruva, Srinivasan Sriram, Jay Lemmon, Josh
  Merel, Greg Wayne, Yuval Tassa, Tom Erez, Ziyun Wang, S.~M.~Ali Eslami,
  Martin~A. Riedmiller, and David Silver.
\newblock Emergence of locomotion behaviours in rich environments.
\newblock {\em ArXiv}, abs/1707.02286, 2017.

\bibitem{Henderson2008HybridRL}
James Henderson, Oliver Lemon, and Kallirroi Georgila.
\newblock Hybrid reinforcement/supervised learning of dialogue policies from
  fixed data sets.
\newblock {\em Computational Linguistics}, 34:487--511, 2008.

\bibitem{Jaques2019WayOB}
Natasha Jaques, Asma Ghandeharioun, Judy~Hanwen Shen, Craig Ferguson, {\`A}gata
  Lapedriza, Noah~J. Jones, Shixiang~Shane Gu, and Rosalind~W. Picard.
\newblock Way off-policy batch deep reinforcement learning of implicit human
  preferences in dialog.
\newblock {\em ArXiv}, abs/1907.00456, 2019.

\bibitem{NetHackOrg}
{Kenneth Lorber}.
\newblock {NetHack Home Page}.
\newblock \url{https://nethack.org}, 2020.
\newblock Accessed: 2020-05-30.

\bibitem{iql}
Ilya Kostrikov, Ashvin Nair, and Sergey Levine.
\newblock Offline reinforcement learning with implicit q-learning.
\newblock {\em ArXiv}, abs/2110.06169, 2021.

\bibitem{cql}
Aviral Kumar, Aurick Zhou, G.~Tucker, and Sergey Levine.
\newblock Conservative q-learning for offline reinforcement learning.
\newblock {\em ArXiv}, abs/2006.04779, 2020.

\bibitem{Kurin2017TheAG}
Vitaly Kurin, Sebastian Nowozin, Katja Hofmann, Lucas Beyer, and B.~Leibe.
\newblock The atari grand challenge dataset.
\newblock {\em ArXiv}, abs/1705.10998, 2017.

\bibitem{nle}
Heinrich Kuttler, Nantas Nardelli, Alexander~H. Miller, Roberta Raileanu, Marco
  Selvatici, Edward Grefenstette, and Tim Rockt{\"a}schel.
\newblock The nethack learning environment.
\newblock {\em ArXiv}, abs/2006.13760, 2020.

\bibitem{Lee2020LearningQL}
Joonho Lee, Jemin Hwangbo, Lorenz Wellhausen, Vladlen Koltun, and Marco Hutter.
\newblock Learning quadrupedal locomotion over challenging terrain.
\newblock {\em Science Robotics}, 5, 2020.

\bibitem{Levine2020OfflineRL}
Sergey Levine, Aviral Kumar, G.~Tucker, and Justin Fu.
\newblock Offline reinforcement learning: Tutorial, review, and perspectives on
  open problems.
\newblock {\em ArXiv}, abs/2005.01643, 2020.

\bibitem{Levine2018LearningHC}
Sergey Levine, Peter Pastor, Alex Krizhevsky, and Deirdre Quillen.
\newblock Learning hand-eye coordination for robotic grasping with deep
  learning and large-scale data collection.
\newblock {\em The International Journal of Robotics Research}, 37:421 -- 436,
  2018.

\bibitem{Li2010ACA}
Lihong Li, Wei Chu, John Langford, and Robert~E. Schapire.
\newblock A contextual-bandit approach to personalized news article
  recommendation.
\newblock {\em ArXiv}, abs/1003.0146, 2010.

\bibitem{Mahler2019LearningAR}
Jeffrey Mahler, Matthew Matl, Vishal Satish, Michael Danielczuk, Bill DeRose,
  Stephen McKinley, and Ken Goldberg.
\newblock Learning ambidextrous robot grasping policies.
\newblock {\em Science Robotics}, 4, 2019.

\bibitem{Mandlekar2021WhatMI}
Ajay Mandlekar, Danfei Xu, J.~Wong, Soroush Nasiriany, Chen Wang, Rohun
  Kulkarni, Li~Fei-Fei, Silvio Savarese, Yuke Zhu, and Roberto Mart'in-Mart'in.
\newblock What matters in learning from offline human demonstrations for robot
  manipulation.
\newblock {\em ArXiv}, abs/2108.03298, 2021.

\bibitem{matthews2022hierarchical}
Michael Matthews, Mikayel Samvelyan, Jack Parker-Holder, Edward Grefenstette,
  and Tim Rockt{\"a}schel.
\newblock Hierarchical kickstarting for skill transfer in reinforcement
  learning, 2022.

\bibitem{moolib2022}
Vegard Mella, Eric Hambro, Danielle Rothermel, and Heinrich K{\"{u}}ttler.
\newblock {moolib: A Platform for Distributed RL}.
\newblock 2022.

\bibitem{Memmesheimer2019SimitateAH}
Raphael Memmesheimer, Ivanna Mykhalchyshyna, Viktor Seib, and Dietrich Paulus.
\newblock Simitate: A hybrid imitation learning benchmark.
\newblock {\em 2019 IEEE/RSJ International Conference on Intelligent Robots and
  Systems (IROS)}, pages 5243--5249, 2019.

\bibitem{atari}
Volodymyr Mnih, Koray Kavukcuoglu, David Silver, Alex Graves, Ioannis
  Antonoglou, Daan Wierstra, and Martin~A. Riedmiller.
\newblock Playing atari with deep reinforcement learning.
\newblock {\em ArXiv}, abs/1312.5602, 2013.

\bibitem{nhwiki}
{NetHack Wiki}.
\newblock {NetHackWiki}.
\newblock \url{https://nethackwiki.com/}, 2020.
\newblock Accessed: 2020-02-01.

\bibitem{OpenAI2019SolvingRC}
OpenAI, Ilge Akkaya, Marcin Andrychowicz, Maciek Chociej, Mateusz Litwin, Bob
  McGrew, Arthur Petron, Alex Paino, Matthias Plappert, Glenn Powell, Raphael
  Ribas, Jonas Schneider, Nikolas~A. Tezak, Jerry Tworek, Peter Welinder,
  Lilian Weng, Qiming Yuan, Wojciech Zaremba, and Lei~M. Zhang.
\newblock Solving rubik's cube with a robot hand.
\newblock {\em ArXiv}, abs/1910.07113, 2019.

\bibitem{paster2022you}
Keiran Paster, Sheila McIlraith, and Jimmy Ba.
\newblock You can't count on luck: Why decision transformers fail in stochastic
  environments.
\newblock {\em arXiv preprint arXiv:2205.15967}, 2022.

\bibitem{petrenko2020sample}
Aleksei Petrenko, Zhehui Huang, Tushar Kumar, Gaurav Sukhatme, and Vladlen
  Koltun.
\newblock Sample factory: Egocentric 3d control from pixels at 100000 fps with
  asynchronous reinforcement learning.
\newblock In {\em International Conference on Machine Learning}, pages
  7652--7662. PMLR, 2020.

\bibitem{Pietquin2011SampleefficientBR}
Olivier Pietquin, Matthieu Geist, Senthilkumar Chandramohan, and Herv{\'e}
  Frezza-Buet.
\newblock Sample-efficient batch reinforcement learning for dialogue management
  optimization.
\newblock {\em ACM Trans. Speech Lang. Process.}, 7:7:1--7:21, 2011.

\bibitem{Prudencio2022ASO}
Rafael~Figueiredo Prudencio, Marcos~R.O.A. Maximo, and Esther~Luna Colombini.
\newblock A survey on offline reinforcement learning: Taxonomy, review, and
  open problems.
\newblock {\em ArXiv}, abs/2203.01387, 2022.

\bibitem{Qin2021NeoRLAN}
Rongjun Qin, Songyi Gao, Xingyuan Zhang, Zhen Xu, Shengkai Huang, Zewen Li,
  Weinan Zhang, and Yang Yu.
\newblock Neorl: A near real-world benchmark for offline reinforcement
  learning.
\newblock {\em ArXiv}, abs/2102.00714, 2021.

\bibitem{Rajeswaran2018LearningCD}
Aravind Rajeswaran, Vikash Kumar, Abhishek Gupta, John Schulman, Emanuel
  Todorov, and Sergey Levine.
\newblock Learning complex dexterous manipulation with deep reinforcement
  learning and demonstrations.
\newblock {\em ArXiv}, abs/1709.10087, 2018.

\bibitem{rlds}
Sabela Ramos, Sertan Girgin, L'eonard Hussenot, Damien Vincent, Hanna
  Yakubovich, Daniel Toyama, Anita Gergely, Piotr Stanczyk, Rapha{\"e}l
  Marinier, Jeremiah Harmsen, Olivier Pietquin, and Nikola Momchev.
\newblock Rlds: an ecosystem to generate, share and use datasets in
  reinforcement learning.
\newblock {\em ArXiv}, abs/2111.02767, 2021.

\bibitem{Ross2011ARO}
St{\'e}phane Ross, Geoffrey~J. Gordon, and J.~Andrew Bagnell.
\newblock A reduction of imitation learning and structured prediction to
  no-regret online learning.
\newblock In {\em AISTATS}, 2011.

\bibitem{Rusu2017SimtoRealRL}
Andrei~A. Rusu, Matej Vecer{\'i}k, Thomas Roth{\"o}rl, Nicolas Manfred~Otto
  Heess, Razvan Pascanu, and Raia Hadsell.
\newblock Sim-to-real robot learning from pixels with progressive nets.
\newblock {\em ArXiv}, abs/1610.04286, 2017.

\bibitem{samvelyan2021minihack}
Mikayel Samvelyan, Robert Kirk, Vitaly Kurin, Jack Parker-Holder, Minqi Jiang,
  Eric Hambro, Fabio Petroni, Heinrich K{\"u}ttler, Edward Grefenstette, and
  Tim Rockt{\"a}schel.
\newblock Minihack the planet: A sandbox for open-ended reinforcement learning
  research.
\newblock {\em arXiv preprint arXiv:2109.13202}, 2021.

\bibitem{kickstarting}
Simon Schmitt, Jonathan~J. Hudson, Augustin Z{\'i}dek, Simon Osindero, Carl
  Doersch, Wojciech~M. Czarnecki, Joel~Z. Leibo, Heinrich K{\"u}ttler, Andrew
  Zisserman, Karen Simonyan, and S.~M.~Ali Eslami.
\newblock Kickstarting deep reinforcement learning.
\newblock {\em ArXiv}, abs/1803.03835, 2018.

\bibitem{schulman2017proximal}
John Schulman, Filip Wolski, Prafulla Dhariwal, Alec Radford, and Oleg Klimov.
\newblock Proximal policy optimization algorithms.
\newblock {\em arXiv preprint arXiv:1707.06347}, 2017.

\bibitem{Sharma2018MultipleIM}
Pratyusha Sharma, Lekha Mohan, Lerrel Pinto, and Abhinav~Kumar Gupta.
\newblock Multiple interactions made easy (mime): Large scale demonstrations
  data for imitation.
\newblock In {\em CoRL}, 2018.

\bibitem{Siegel2020KeepDW}
Noah Siegel, Jost~Tobias Springenberg, Felix Berkenkamp, Abbas Abdolmaleki,
  Michael Neunert, Thomas Lampe, Roland Hafner, and Martin~A. Riedmiller.
\newblock Keep doing what worked: Behavioral modelling priors for offline
  reinforcement learning.
\newblock {\em ArXiv}, abs/2002.08396, 2020.

\bibitem{alphago}
David Silver, Aja Huang, Chris~J Maddison, Arthur Guez, Laurent Sifre, George
  Van Den~Driessche, Julian Schrittwieser, Ioannis Antonoglou, Veda
  Panneershelvam, Marc Lanctot, et~al.
\newblock Mastering the game of go with deep neural networks and tree search.
\newblock {\em nature}, 529(7587):484--489, 2016.

\bibitem{Stadie2017ThirdPersonIL}
Bradly~C. Stadie, P.~Abbeel, and Ilya Sutskever.
\newblock Third-person imitation learning.
\newblock {\em ArXiv}, abs/1703.01703, 2017.

\bibitem{Strehl2010LearningFL}
Alexander~L. Strehl, John Langford, Lihong Li, and Sham~M. Kakade.
\newblock Learning from logged implicit exploration data.
\newblock In {\em NIPS}, 2010.

\bibitem{d3rlpy}
Michita~Imai Takuma~Seno.
\newblock d3rlpy: An offline deep reinforcement library.
\newblock In {\em NeurIPS 2021 Offline Reinforcement Learning Workshop},
  December 2021.

\bibitem{telegraph}
The Telegraph.
\newblock The 25 hardest video games ever.
\newblock 2021.
\newblock Accessed: 2021-05-05.

\bibitem{Thomas2017PredictiveOP}
Philip~S. Thomas, Georgios Theocharous, Mohammad Ghavamzadeh, Ishan Durugkar,
  and Emma Brunskill.
\newblock Predictive off-policy policy evaluation for nonstationary decision
  problems, with applications to digital marketing.
\newblock In {\em AAAI}, 2017.

\bibitem{bco}
Faraz Torabi, Garrett Warnell, and Peter Stone.
\newblock Behavioral cloning from observation.
\newblock {\em ArXiv}, abs/1805.01954, 2018.

\bibitem{Toyer2020TheMB}
Sam Toyer, Rohin Shah, Andrew Critch, and Stuart~J. Russell.
\newblock The magical benchmark for robust imitation.
\newblock {\em ArXiv}, abs/2011.00401, 2020.

\bibitem{alphastar}
Oriol Vinyals, Igor Babuschkin, Wojciech~M. Czarnecki, Micha{\"e}l Mathieu,
  Andrew Dudzik, Junyoung Chung, David~H. Choi, Richard Powell, Timo Ewalds,
  Petko Georgiev, Junhyuk Oh, Dan Horgan, Manuel Kroiss, Ivo Danihelka, Aja
  Huang, L.~Sifre, Trevor Cai, John~P. Agapiou, Max Jaderberg, Alexander~Sasha
  Vezhnevets, R{\'e}mi Leblond, Tobias Pohlen, Valentin Dalibard, David Budden,
  Yury Sulsky, James Molloy, Tom~Le Paine, Caglar Gulcehre, Ziyun Wang, Tobias
  Pfaff, Yuhuai Wu, Roman Ring, Dani Yogatama, Dario W{\"u}nsch, Katrina
  McKinney, Oliver Smith, Tom Schaul, Timothy~P. Lillicrap, Koray Kavukcuoglu,
  Demis Hassabis, Chris Apps, and David Silver.
\newblock Grandmaster level in starcraft ii using multi-agent reinforcement
  learning.
\newblock {\em Nature}, pages 1--5, 2019.

\bibitem{starcraft2}
Oriol Vinyals, Timo Ewalds, Sergey Bartunov, Petko Georgiev, Alexander~Sasha
  Vezhnevets, Michelle Yeo, Alireza Makhzani, Heinrich K{\"u}ttler, John~P.
  Agapiou, Julian Schrittwieser, John Quan, Stephen Gaffney, Stig Petersen,
  Karen Simonyan, Tom Schaul, H.~V. Hasselt, David Silver, Timothy~P.
  Lillicrap, Kevin Calderone, Paul Keet, Anthony Brunasso, David Lawrence,
  Anders Ekermo, Jacob Repp, and Rodney Tsing.
\newblock Starcraft ii: A new challenge for reinforcement learning.
\newblock {\em ArXiv}, abs/1708.04782, 2017.

\bibitem{brac}
Yifan Wu, G.~Tucker, and Ofir Nachum.
\newblock Behavior regularized offline reinforcement learning.
\newblock {\em ArXiv}, abs/1911.11361, 2019.

\bibitem{brac+}
Chi Zhang, Sanmukh~Rao Kuppannagari, and Viktor~K. Prasanna.
\newblock Brac+: Improved behavior regularized actor critic for offline
  reinforcement learning.
\newblock In {\em ACML}, 2021.

\bibitem{rlunplugged}
Çaglar G{\"u}lçehre, Ziyun Wang, Alexander Novikov, Tom~Le Paine,
  Sergio~Gomez Colmenarejo, Konrad Zolna, Rishabh Agarwal, Josh Merel,
  Daniel~Jaymin Mankowitz, Cosmin Paduraru, Gabriel Dulac-Arnold, Jerry~Zheng
  Li, Mohammad Norouzi, Matthew~D. Hoffman, Nicolas Manfred~Otto Heess, and
  Nando de~Freitas.
\newblock Rl unplugged: A collection of benchmarks for offline reinforcement
  learning.
\newblock In {\em NeurIPS}, 2020.

\end{thebibliography}

\newpage
\section{Checklist}
\begin{enumerate}

\item For all authors...
\begin{enumerate}
  \item Do the main claims made in the abstract and introduction accurately reflect the paper's contributions and scope?
     \answerYes{}
  \item Did you describe the limitations of your work?
    \answerYes{} See Appendix~\ref{app:limitations}. 
  \item Did you discuss any potential negative societal impacts of your work?
    \answerYes{}. See Appendix~\ref{app:broader-impact}.
  \item Have you read the ethics review guidelines and ensured that your paper conforms to them?
    \answerYes{}
\end{enumerate}

\item If you are including theoretical results...
\begin{enumerate}
  \item Did you state the full set of assumptions of all theoretical results?
    \answerNA{}
	\item Did you include complete proofs of all theoretical results?
    \answerNA{}
\end{enumerate}

\item If you ran experiments (e.g. for benchmarks)...
\begin{enumerate}
  \item Did you include the code, data, and instructions needed to reproduce the main experimental results (either in the supplemental material or as a URL)?
      \answerYes{} Code will be supplied with the dataset at the link provided in Appendix \ref{sec:access}.
  \item Did you specify all the training details (e.g., data splits, hyperparameters, how they were chosen)?
    \answerYes{} See Appendix \ref{sec:appendix-experiments}
	\item Did you report error bars (e.g., with respect to the random seed after running experiments multiple times)?
    \answerYes{} We report standard deviations across seeds for all results.
	\item Did you include the total amount of compute and the type of resources used (e.g., type of GPUs, internal cluster, or cloud provider)?
    \answerYes{}  See Appendix \ref{sec:appendix-experiments}
\end{enumerate}

\item If you are using existing assets (e.g., code, data, models) or curating/releasing new assets...
\begin{enumerate}
  \item If your work uses existing assets, did you cite the creators?
    \answerYes{} Although we do not use ``assets" per se, we do cite the bot AutoAscend by linking to it and citing the author's own write up in the NetHack Challenge report.\cite{hambro2022insights}
  \item Did you mention the license of the assets?
     \answerYes{} We mention the license of the dataset (same as \NLE{}) in Appendix \ref{sec:license}
  \item Did you include any new assets either in the supplemental material or as a URL?
    \answerYes{} This is a dataset paper, so the `assets' are the data, which is of course linked to in Appendix~\ref{sec:access}.
  \item Did you discuss whether and how consent was obtained from people whose data you're using/curating?
  \answerNA{} We do not discuss this in the paper because: 1) some of the data is from an open source bot and 2) the human data is already public and completely anonymous. We did  discuss this with the site creator (who collected the data, and put up terms of use) and we obtained his consent, as well as the consent of the creators of the bot.
  \item Did you discuss whether the data you are using/curating contains personally identifiable information or offensive content?
    \answerYes{} We discuss in Appendix \ref{sec:filtering}
\end{enumerate}

\item If you used crowdsourcing or conducted research with human subjects...
\begin{enumerate}
  \item Did you include the full text of instructions given to participants and screenshots, if applicable?
    \answerNA{}
  \item Did you describe any potential participant risks, with links to Institutional Review Board (IRB) approvals, if applicable?
    \answerNA{}
  \item Did you include the estimated hourly wage paid to participants and the total amount spent on participant compensation?
    \answerNA{}
\end{enumerate}

\end{enumerate}

\newpage

\section{Datasets Checklist}
Include extra information in the appendix. This section will often be part of the supplemental material. Please see the call on the NeurIPS website for links to additional guides on dataset publication.

\begin{enumerate}

\item Submission introducing new datasets must include the following in the supplementary materials:
\begin{enumerate}
  \item Dataset documentation and intended uses. Recommended documentation frameworks include datasheets for datasets, dataset nutrition labels, data statements for NLP, and accountability frameworks.  \replychecklist[We document the code to use the dataset in the repository and provide many references in the Appendices.]{}
  \item URL to website/platform where the dataset/benchmark can be viewed and downloaded by the reviewers.  \replychecklist[We provide this in Appendix \ref{sec:access}]{}
  \item Author statement that they bear all responsibility in case of violation of rights, etc., and confirmation of the data license. \replychecklist[We provide this in Appendix \ref{sec:license} and upon submission to OpenReview]{}
  \item Hosting, licensing, and maintenance plan. The choice of hosting platform is yours, as long as you ensure access to the data (possibly through a curated interface) and will provide the necessary maintenance. \replychecklist[We provide this in Appendix \ref{sec:access}]{}
\end{enumerate}

\item To ensure accessibility, the supplementary materials for datasets must include the following:
\begin{enumerate}
  \item Links to access the dataset and its metadata. This can be hidden upon submission if the dataset is not yet publicly available but must be added in the camera-ready version. In select cases, e.g when the data can only be released at a later date, this can be added afterward. Simulation environments should link to (open source) code repositories. 
  \replychecklist[We provide this in Appendix \ref{sec:access}]{}
  \item The dataset itself should ideally use an open and widely used data format. Provide a detailed explanation on how the dataset can be read. For simulation environments, use existing frameworks or explain how they can be used.
  \replychecklist[We provide example usage, documentation, and detailed specification in how data is stored in several Appendices]{}
  \item Long-term preservation: It must be clear that the dataset will be available for a long time, either by uploading to a data repository or by explaining how the authors themselves will ensure this.
    \replychecklist[We provide a maintenance plan in Appendix \ref{sec:access}]{}
  \item Explicit license: Authors must choose a license, ideally a CC license for datasets, or an open source license for code (e.g. RL environments).
  \replychecklist[We provide an open license in Appendix \ref{sec:license}]{}
  \item Add structured metadata to a dataset's meta-data page using Web standards (like schema.org and DCAT): This allows it to be discovered and organized by anyone. If you use an existing data repository, this is often done automatically.  \replychecklist[We are in the process of building a site to host, but are happy to do this when completed.]{}
  \item Highly recommended: a persistent dereferenceable identifier (e.g. a DOI minted by a data repository or a prefix on identifiers.org) for datasets, or a code repository (e.g. GitHub, GitLab,...) for code. If this is not possible or useful, please explain why. \replychecklist[Code and instructions will always live on GitHub]{}.
\end{enumerate}

\item For benchmarks, the supplementary materials must ensure that all results are easily reproducible. Where possible, use a reproducibility framework such as the ML reproducibility checklist, or otherwise guarantee that all results can be easily reproduced, i.e. all necessary datasets, code, and evaluation procedures must be accessible and documented.
\replychecklist[We provide access to code for experiments and will tidy and rehost upon release of the dataset.]{}.

\item For papers introducing best practices in creating or curating datasets and benchmarks, the above supplementary materials are not required.
\end{enumerate}

\newpage
\appendix

\section{Accessing the Dataset}
\label{sec:access}

For the duration of the review the information on how to download, install and use \NLD{}  will be available at \url{https://github.com/dungeonsdatasubmission/dungeonsdata-neurips2022}. This link contains a \texttt{README.md} file with instructions on how to download zip files for \NLDact{} and \NLDobs{}, as well as instructions on how to install and use the \DatasetObject{}, and includes code used in the running of experiment. 

After review, this information will migrate to the \texttt{README.md} for the \NLE{} repo, hosted at \url{https://github.com/facebookresearch/nle}.

\paragraph{Hosting \& Maintenance Plan}
The dataset is hosted in a dedicated, publicly accessible Facebook adminstered S3 bucket. This should be permanently available for the foreseeable future, and full links to download are provided in the \NLE{} repo.

\paragraph{\NLDact{}}
\begin{itemize}
     \item \texttt{https://dl.fbaipublicfiles.com/nld/nld-aa/nld-aa-dir-aa.zip}
     \item \texttt{https://dl.fbaipublicfiles.com/nld/nld-aa/nld-aa-dir-ab.zip}
     \item \texttt{https://dl.fbaipublicfiles.com/nld/nld-aa/nld-aa-dir-ac.zip}
     \item \texttt{https://dl.fbaipublicfiles.com/nld/nld-aa/nld-aa-dir-ad.zip}
     \item \texttt{https://dl.fbaipublicfiles.com/nld/nld-aa/nld-aa-dir-ae.zip}
     \item \texttt{https://dl.fbaipublicfiles.com/nld/nld-aa/nld-aa-dir-af.zip}
     \item \texttt{https://dl.fbaipublicfiles.com/nld/nld-aa/nld-aa-dir-ag.zip}
     \item \texttt{https://dl.fbaipublicfiles.com/nld/nld-aa/nld-aa-dir-ah.zip}
     \item \texttt{https://dl.fbaipublicfiles.com/nld/nld-aa/nld-aa-dir-ai.zip}
     \item \texttt{https://dl.fbaipublicfiles.com/nld/nld-aa/nld-aa-dir-aj.zip}
     \item \texttt{https://dl.fbaipublicfiles.com/nld/nld-aa/nld-aa-dir-ak.zip}
     \item \texttt{https://dl.fbaipublicfiles.com/nld/nld-aa/nld-aa-dir-al.zip}
     \item \texttt{https://dl.fbaipublicfiles.com/nld/nld-aa/nld-aa-dir-am.zip}
     \item \texttt{https://dl.fbaipublicfiles.com/nld/nld-aa/nld-aa-dir-an.zip}
     \item \texttt{https://dl.fbaipublicfiles.com/nld/nld-aa/nld-aa-dir-ao.zip}
     \item \texttt{https://dl.fbaipublicfiles.com/nld/nld-aa/nld-aa-dir-ap.zip}
\end{itemize}

\paragraph{\NLDobs{}}
\begin{itemize}
    \item \texttt{https://dl.fbaipublicfiles.com/nld/nld-nao/nld-nao-dir-aa.zip}
    \item \texttt{https://dl.fbaipublicfiles.com/nld/nld-nao/nld-nao-dir-ab.zip}
    \item \texttt{https://dl.fbaipublicfiles.com/nld/nld-nao/nld-nao-dir-ac.zip}
    \item \texttt{https://dl.fbaipublicfiles.com/nld/nld-nao/nld-nao-dir-ad.zip}
    \item \texttt{https://dl.fbaipublicfiles.com/nld/nld-nao/nld-nao-dir-ae.zip}
    \item \texttt{https://dl.fbaipublicfiles.com/nld/nld-nao/nld-nao-dir-af.zip}
    \item \texttt{https://dl.fbaipublicfiles.com/nld/nld-nao/nld-nao-dir-ag.zip}
    \item \texttt{https://dl.fbaipublicfiles.com/nld/nld-nao/nld-nao-dir-ah.zip}
    \item \texttt{https://dl.fbaipublicfiles.com/nld/nld-nao/nld-nao-dir-ai.zip}
    \item \texttt{https://dl.fbaipublicfiles.com/nld/nld-nao/nld-nao-dir-aj.zip}
    \item \texttt{https://dl.fbaipublicfiles.com/nld/nld-nao/nld-nao-dir-ak.zip}
    \item \texttt{https://dl.fbaipublicfiles.com/nld/nld-nao/nld-nao-dir-al.zip}
    \item \texttt{https://dl.fbaipublicfiles.com/nld/nld-nao/nld-nao-dir-am.zip}
    \item \texttt{https://dl.fbaipublicfiles.com/nld/nld-nao/nld-nao-dir-an.zip}
    \item \texttt{https://dl.fbaipublicfiles.com/nld/nld-nao/nld-nao-dir-ao.zip}
    \item \texttt{https://dl.fbaipublicfiles.com/nld/nld-nao/nld-nao-dir-ap.zip}
    \item \texttt{https://dl.fbaipublicfiles.com/nld/nld-nao/nld-nao-dir-aq.zip}
    \item \texttt{https://dl.fbaipublicfiles.com/nld/nld-nao/nld-nao-dir-ar.zip}
    \item \texttt{https://dl.fbaipublicfiles.com/nld/nld-nao/nld-nao-dir-as.zip}
    \item \texttt{https://dl.fbaipublicfiles.com/nld/nld-nao/nld-nao-dir-at.zip}
    \item \texttt{https://dl.fbaipublicfiles.com/nld/nld-nao/nld-nao-dir-au.zip}
    \item \texttt{https://dl.fbaipublicfiles.com/nld/nld-nao/nld-nao-dir-av.zip}
    \item \texttt{https://dl.fbaipublicfiles.com/nld/nld-nao/nld-nao-dir-aw.zip}
    \item \texttt{https://dl.fbaipublicfiles.com/nld/nld-nao/nld-nao-dir-ax.zip}
    \item \texttt{https://dl.fbaipublicfiles.com/nld/nld-nao/nld-nao-dir-ay.zip}
    \item \texttt{https://dl.fbaipublicfiles.com/nld/nld-nao/nld-nao-dir-az.zip}
    \item \texttt{https://dl.fbaipublicfiles.com/nld/nld-nao/nld-nao-dir-ba.zip}
    \item \texttt{https://dl.fbaipublicfiles.com/nld/nld-nao/nld-nao-dir-bb.zip}
    \item \texttt{https://dl.fbaipublicfiles.com/nld/nld-nao/nld-nao-dir-bc.zip}
    \item \texttt{https://dl.fbaipublicfiles.com/nld/nld-nao/nld-nao-dir-bd.zip}
    \item \texttt{https://dl.fbaipublicfiles.com/nld/nld-nao/nld-nao-dir-be.zip}
    \item \texttt{https://dl.fbaipublicfiles.com/nld/nld-nao/nld-nao-dir-bf.zip}
    \item \texttt{https://dl.fbaipublicfiles.com/nld/nld-nao/nld-nao-dir-bg.zip}
    \item \texttt{https://dl.fbaipublicfiles.com/nld/nld-nao/nld-nao-dir-bh.zip}
    \item \texttt{https://dl.fbaipublicfiles.com/nld/nld-nao/nld-nao-dir-bi.zip}
    \item \texttt{https://dl.fbaipublicfiles.com/nld/nld-nao/nld-nao-dir-bj.zip}
    \item \texttt{https://dl.fbaipublicfiles.com/nld/nld-nao/nld-nao-dir-bk.zip}
    \item \texttt{https://dl.fbaipublicfiles.com/nld/nld-nao/nld-nao-dir-bl.zip}
    \item \texttt{https://dl.fbaipublicfiles.com/nld/nld-nao/nld-nao-dir-bm.zip}
    \item \texttt{https://dl.fbaipublicfiles.com/nld/nld-nao/nld-nao-dir-bn.zip}
    \item \texttt{https://dl.fbaipublicfiles.com/nld/nld-nao/nld-nao\_xlogfiles.zip}
\end{itemize}

\section{License}
\label{sec:license}

Data is provided under the NetHack General Public License - A GPL style license that is used to covered the NetHack Game since 1989. The license can be found at \url{https://github.com/facebookresearch/nle/blob/main/LICENSE}, and is used by both the official NetHack repository\footnote{\url{https://github.com/NetHack/NetHack}} and \NLE{}. 

\section{\NLE{} \version{0}{9}{0}}
\label{sec:nle_version}

\NLE{} \version{0}{9}{0} will be the latest version of \NLE{} released, and will be released with \NLD{}. The release branch is will be cut from \texttt{main} branch on the repo, hosted at (\url{https://github.com/facebookresearch/nle/}), and deployed to PyPI. 
The branch currently holds all the changes needed for compatibility with \NLD{}, including the \DatasetObject{}. A full changelog will be made public after release but the changes include:

\begin{itemize}
    \item New ability to record \ttyrecthreebz{} files directly from \NLE{}. Previous versions recorded a modified \ttyrecbz{}.
    \item The additional logging to an \xlogfile{} for each episodes that finish naturally (\ie{} when \texttt{done} is \texttt{True}).
    \item A C++ \texttt{Converter} class to load \ttyrecbz{}/\ttyrecthreebz{} files directly in to NumPy arrays, based on a V100 Terminal emulator (from \texttt{libtmt}\footnote{\url{https://github.com/deadpixi/libtmt}}).
    \item The \DatasetObject{} class, to marshall the \texttt{Converter} objects into an \texttt{Torch.IterableDataset} interface, and handle metadata.
\end{itemize}

\section{The \ttyrec{} format}
\label{sec:ttyrec}

The \ttyrec{} is a file format that has historically been used to store recordings of terminal-based NetHack games.\footnote{\url{https://nethackwiki.com/wiki/Ttyrec}}
The format consists of a stack of \textit{frames} where each \textit{frame} consists of a 12-byte header immediately followed by a variable-size buffer of terminal instructions. 
The header contains 8 bytes of time information (storing \textit{when} these updates were recorded) and 4 bytes indicating the size of the subsequent buffer (storing \textit{what} these updates were). 
The buffer's contents are then the instructions to be fed to a terminal that will be rendering \ttyrec{}. As such, this buffer will then generally consist of terminal-specific Escape Sequences\footnote{\url{https://en.wikipedia.org/wiki/Escape\_sequence}} and text to be printed.

\NLE{}'s \ttyrecthree{} has a 13-byte header, adding 1-byte channel to indicate \textit{what kind} of content is in the buffer: \texttt{0} for terminal output; \texttt{1} for terminal input (i.e the keypresses corresponding to actions in the game); and, \texttt{2} for in-game score. These latter two channels are written just after and just before an action is taken (respectively).

It is worth noting that \textit{frames} with channel \texttt{0} are written whenever NetHack intends to write to the screen, and therefore several may written before an action is required. An example of such an event occurs when zapping wand which animates a beam from the player. Such animations may result in a stack of frames with channels that resemble something like \texttt{[0 2 1 0 0 0 2 1 ...]}, where several different states might correspond to one action. In this case \ttyrecthreebz{} rendering loads  the final state of the of the terminal NumPy arrays, alongside the score and action, while \ttyrecbz{} rendering loads every state/frame, having no clear delineation of where actions take place.

\begin{figure}
    \centering
    \includegraphics[width=0.9\linewidth]{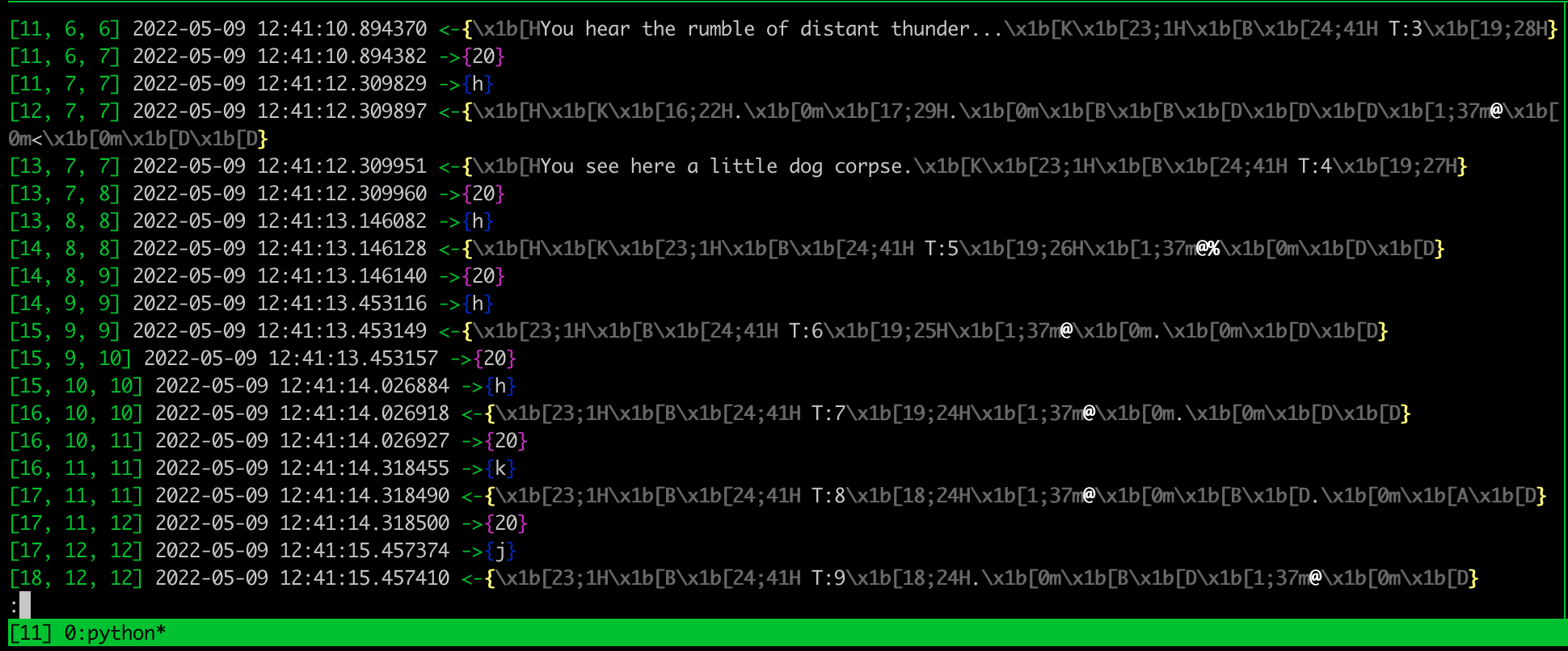}
    \caption{A screenshot of a tool to read raw \ttyrecbz{}/\ttyrecthreebz{} files, included in \NLE{} \version{0}{9}{0}. The numbers in green represent the frame count for each channel [0, 1, 2], followed by the timestamp, and the buffer contents in braces: channel \texttt{0} is displayed in yellow braces, \texttt{1} in blue braces, and \texttt{2} in pink braces. Terminal Escape Sequences are printed in dark grey.}
    \label{fig:readtty}
\end{figure}

To further inspect \ttyrecbz{} and \ttyrecthreebz{}, \NLE{} \version{0}{9}{0} comes with an adapted \texttt{read\_tty.py} script that displays the raw, decompressed \ttyrec{}/\ttyrecthree{} contents, shown in Figure \ref{fig:readtty}.

\section{\NLD{} Gameplay Metadata Reference}
\label{sec:metadata}

The gameplay metadata in \NLD{} is identical to that stored NetHacks's \xlogfile{} at the end of a game, with the addition of a \textit{gameid} which is used by \NLD{} to identify games.  The metadata consists of:

\begin{enumerate}[itemsep=0.05em]
  
    \item{} \textit{gameid} \texttt{(int)} - A unique id for the game, created by the local database.  
    \item{} \textit{version} \texttt{(str)} - The version of NetHack played.
    \item{} \textit{points} \texttt{(int)} - The final in-game score of the episode.
    \item{} \textit{deathdnum} \texttt{(int)} - The dungeon number where you died. These correspond to the dungeons found in \texttt{dungeon.def} file:
    \begin{itemize}
        \item \texttt{0} -  The Dungeons of Doom
        \item \texttt{1} -  Gehennom
        \item \texttt{2} -  The Gnomish Mines
        \item \texttt{3} -  The Quest
        \item \texttt{4} -  Sokoban
        \item \texttt{5} -  Fort Ludios
        \item \texttt{6} -  Elemental Planes
    \end{itemize}

    \item{} \textit{deathlev} \texttt{(int)} - The dungeon level where you died.\footnote{For more, see \url{https://nethackwiki.com/wiki/Mazes_of_Menace}}
    \item{} \textit{maxlvl} \texttt{(int)} - The deepest dungeon level  you reached.
    \item{} \textit{hp} \texttt{(int)} - The number of hit points you ended the game with.
    \item{} \textit{maxhp} \texttt{(int)} - The number of hit points you'd have had  at ``full health'' at the game end.
    \item{} \textit{deaths} \texttt{(int)} - The number of times you died.\footnote{In rare cases, you can die more than once. See \url{https://nethackwiki.com/wiki/Amulet_of_life_saving}}
    \item{} \textit{deathdate} \texttt{(int)} - The day the game ended (format \texttt{YYYYMMDD}).
    \item{} \textit{birthdate} \texttt{(int)} - The day the game started (format \texttt{YYYYMMDD}).
    \item{} \textit{uid} \texttt{(int)} - An id used (in conjunction with player name) to identify players (for save games etc).
    \item{} \textit{role} \texttt{(str)} - The Role of the player. The 13 roles are: 
    \begin{itemize}
        \item \texttt{Arc} - Archaelogist
        \item \texttt{Bar} - Barbarian
        \item \texttt{Cav} - Cave(wo)man
        \item \texttt{Hea} - Healer
        \item \texttt{Kni} - Knight
        \item \texttt{Mon} - Monk
        \item \texttt{Pri} - Priest(ess)
        \item \texttt{Ran} - Ranger
        \item \texttt{Rog} - Rogue
        \item \texttt{Sam} - Samurai
        \item \texttt{Tou} - Tourist
        \item \texttt{Val} - Valkyrie
        \item \texttt{Wiz} - Wizard
    \end{itemize}
    \item{} \textit{race}  \texttt{(str)} - The Race of the player. The 5 races are: 
    \begin{itemize}
        \item \texttt{Dwa} - Dwarf
        \item \texttt{Elf} - Elf
        \item \texttt{Gno} - Gnome
        \item \texttt{Hum} - Human
        \item \texttt{Orc} - Orc
    \end{itemize}
    \item{} \textit{gender} \texttt{(str)} - The Gender of the player. The 2 genders are: 
    \begin{itemize}
        \item \texttt{Fem} - Female
        \item \texttt{Mal} - Male
    \end{itemize}
    \item{} \textit{align} \texttt{(str)} - The Alignment of the player. The 3 alignments are:
    \begin{itemize}
        \item \texttt{Cha} - Chaotic
        \item \texttt{Law} - Lawful
        \item \texttt{Neu} - Neutral
    \end{itemize}
    \item{} \textit{name} \texttt{(str)} - The name of the player. These are pseudonymised in the database for \NLDobs{}.
    \item{} \textit{death} \texttt{(str)} - A description of the manner in which the player died or ended the game.
    \item{} \textit{conduct} \texttt{(str)} - a bitfield encoding the conducts\footnote{\url{https://nethackwiki.com/wiki/Conduct}} completed in the game. The bitfield encodes:
    \begin{itemize}
        \item \texttt{0x001}: Foodless
        \item \texttt{0x002}: Vegan
        \item \texttt{0x004}: Vegetarian
        \item \texttt{0x008}: Atheist
        \item \texttt{0x010}: Weaponless
        \item \texttt{0x020}: Pacifist
        \item \texttt{0x040}: Illiterate
        \item \texttt{0x080}: Polypileless
        \item \texttt{0x100}: Polyselfless
        \item \texttt{0x200}: Wishless
        \item \texttt{0x400}: Artifact wishless
        \item \texttt{0x800}: Genocideless
    \end{itemize}
    \item{} \textit{turns} \texttt{(int)} - The number of in-game turns played by the player. This may not correspond to transitions, as several moves do not advance the in-game clock (eg checking the inventory or moving into a wall).
    \item{} \textit{achieve} \texttt{(str)} -  a bitfield encoding the achievements\footnote{\url{https://nethackwiki.com/wiki/Xlogfile}} attained in the game. The bitfield encodes:
    \begin{itemize}
        \item \texttt{0x0001}: Got the Bell of Opening
        \item \texttt{0x0002}: Entered Gehennom
        \item \texttt{0x0004}: Got the Candelabrum of Invocation
        \item \texttt{0x0008}: Got the Book of the Dead
        \item \texttt{0x0010}: Performed the Invocation
        \item \texttt{0x0020}: Got the Amulet of Yendor
        \item \texttt{0x0040}: Was in the End Game 
        \item \texttt{0x0080}: Was on the Astral Plane 
        \item \texttt{0x0100}: Ascended 
        \item \texttt{0x0200}: Got the Luckstone at Mines’ End
        \item \texttt{0x0400}: Finished Sokoban 
        \item \texttt{0x0800}: Killed Medusa 
        \item \texttt{0x1000}: Zen conduct intact 
        \item \texttt{0x2000}: Nudist conduct intact
    \end{itemize}
    \item{} \textit{realtime} \texttt{(int)} - the duration of the game in seconds
    \item{} \textit{starttime} \texttt{(int)} - the start time of the game as an (epoch) unix timestamp.
    \item{} \textit{endtime} \texttt{(int)} - the end time of the game as an (epoch) unix timestamp.
    \item{} \textit{gender0} \texttt{(str)} - the starting gender of the player. Same options as  \textit{gender}.
    \item{} \textit{align0} \texttt{(str)} - the starting alignment of the player. Same options as \textit{align}.
    \item{} \textit{flags} \texttt{(str)} - a bitfield encoding some additional state of the game. The bitfield encodes: \begin{itemize}
        \item \texttt{0x1}: Wizard mode
        \item \texttt{0x2}: Discover mode
        \item \texttt{0x4}: Never loaded a Bones file
    \end{itemize}

\end{enumerate}

\section{\NLD{} Observations Reference}
\label{sec:observation}

The \DatasetObject{} loads minibatches of observations, with batch size B, and sequence/unroll length T. The following observations are present in each minibatch dictionary:

\begin{tabular}{l c c p{7.5cm}}
     \toprule
     Name &  Type & Shape &  Description\\
     \midrule
    \texttt{tty\_chars} & \texttt{np.uint8} & [B, T, H, W] & The on-screen characters (default screen size 80 x 24).\\
    \texttt{tty\_colors} & \texttt{np.int8} & [B, T, H, W] & The on-screen colors for each character. \\
    \texttt{tty\_cursor} & \texttt{np.int16} & [B, T, 2] & The coordinates of the on-screen cursor.\\
    \texttt{timestamps} & \texttt{np.int64} & [B, T] & The time each frame was recorded.\\
    \texttt{gameids} & \texttt{np.int32} & [B, T] & The \textit{gameid} for the episode being rendered.\\
    \texttt{done} & \texttt{np.uint8} & [B, T] & An indicator that this frame is from a new \textit{gameid}.\\
    \texttt{scores}* & \texttt{np.int32} & [B, T] & The in-game score at this time (\ie{} before keypress).\\
    \texttt{keypresses}* & \texttt{np.uint8} & [B, T] & The keypress the player made in response to this state (\ie{} after seeing \texttt{tty\_*}, \texttt{scores}, etc). \\

    \bottomrule
\end{tabular}

The observations marked with an asterisk (*) are only available in \ttyrecthreebz{} datasets. 
\section{\DatasetObject{} Reference}
\label{sec:api}

\subsection{Design Overview} 
To load minibatches, the \DatasetObject{} performs two main actions: first, at construction time, it obtains the list of gameids and \ttyrecbz{}/\ttyrecthreebz{} filepaths that constitute dataset, and second, at usage time, it constructs an Python generator\footnote{\url{https://peps.python.org/pep-0255/}} that will sequentially load these files into fixed NumPy arrays. 

The first of these actions is performed by querying a local \textit{sqlite3} database file, which acts as the metadata information hub for all datasets. This database file need only be populated once, and from thenceforth can be copied, shared or swapped out, without issue.  For more on how to add datasets see the Tutorial Notebook\footnote{\url{https://github.com/dungeonsdatasubmission/dungeonsdata-neurips2022/blob/main/usage_tutorial_notebook.ipynb}}, or Appendix \ref{sec:appendix-populate}. The metadata in this database includes the location of files for each game, the association of which games are in which dataset, and the gameplay metadata for each game (see Appendix \ref{sec:metadata}). For more on the database schema see Appendix \ref{appendix:schema}.

The second of these actions is performed by the \DatasetObject{} itself, by creating a lightweight C++ \texttt{Converter} object for each batch index in a generator, and loading from these objects when \texttt{next} is called. This process can be modified in a variety of different ways, including looping forever, shuffling, or using a threadpool. For more on the way these can be modified see Appendix \ref{sec:appendix-api}.

\subsection{Database Layout}
\label{appendix:schema}
Table \ref{tab:dbschema} shows the layout of the database, where fields of the same name map directly to each other. Each dataset is associated with a string (dataset\_name), and once populated the dataset can be retrieved with this string.  

\begin{table}[]
    \caption{The layout of the \textit{sqlite3} database used by the \DatasetObject{}. The \texttt{roots} table stores information about the root directory of the files in a dataset, as well as \ttyrec{} version, while \texttt{ttyrecs} table stores the data about the files and what games they belong to. The \texttt{datasets} table then associates the gameids to their dataset, and \texttt{games} contains the gameplay metadata for each gameid.}
    \centering
        \begin{tabular}{ll}
        \toprule
        \texttt{roots} & \\
        \midrule
        dataset\_name   & TEXT \\
        root            & TEXT \\
        ttyrec\_version & INTEGER\\
        \bottomrule
        \\
        \\
        \\
        \toprule
        \texttt{datasets} & \\
        \midrule
        gameids & INTEGER\\
        dataset\_name & TEXT\\
        \bottomrule
        \\
        \\
        \\
        \toprule
        \texttt{ttyrecs} & \\
        \midrule
            path      &  TEXT\\
            part      &  INTEGER\\
            size      &  INTEGER\\
            mtime     &  REAL\\
            gameid    &  INTEGER\\
        \bottomrule
        \\
        \\
        \\
        \toprule
        \texttt{meta} & \\
        \midrule
            ctime      &  REAL\\
            mtime      &  REAL\\
        \bottomrule
        \end{tabular}
        \quad
        \begin{tabular}{ll}
        \toprule
        \texttt{games} & \\
        \midrule
            gameid     & INTEGER \\
            version    & TEXT\\
            points  &    INTEGER\\
            deathdnum &  INTEGER\\
            deathlev   & INTEGER\\
            maxlvl  &    INTEGER\\
            hp      &    INTEGER\\
            maxhp   &    INTEGER\\
            deaths  &    INTEGER\\
            deathdate&   INTEGER\\
            birthdate&   INTEGER\\
            uid     &    INTEGER\\
            role    &    TEXT\\
            race    &    TEXT\\
            gender  &    TEXT\\
            align   &    TEXT\\
            name    &    TEXT\\
            death   &    TEXT\\
            conduct &    TEXT\\
            turns   &    INTEGER\\
            achieve &    TEXT\\
            realtime&    INTEGER\\
            starttime&   INTEGER\\
            endtime &    INTEGER\\
            gender0 &    TEXT\\
            align0  &    TEXT\\
            flags   &    TEXT\\
        \bottomrule
        \end{tabular}

    \label{tab:dbschema}
\end{table}

\subsection{\DatasetObject{} API}
\label{sec:appendix-api}
The \DatasetObject{} is documented primarily through Python docstrings, which provide a helpful way automatically documenting code objects, that can be queried by using Python's \texttt{help} function. An example usage can be clearly seen in the tutorial notebook\footnote{\url{https://github.com/dungeonsdatasubmission/dungeonsdata-neurips2022/blob/main/usage_tutorial_notebook.ipynb}} supplied with the submission, that will also migrate to \NLE{}.

The \DatasetObject{} accepts the following arguments, with all but first taking reasonable default values:
\begin{enumerate}[itemsep=0.05em]
    \item \texttt{dataset\_name} - \texttt{(str)} - The name of the dataset to load. Each datasets is associated with a unique name. 
    \item \texttt{batch\_size} - \texttt{(int)} - The number of simulataneous episodes to load (\ie{} size B).
    \item \texttt{seq\_length} - \texttt{(int)} - The number of steps to unroll for each episode in a minibatch (\ie{} size T).
    \item \texttt{rows} - \texttt{(int)} - The row size of the terminal to emulate (\ie{} size H).
    \item \texttt{cols} - \texttt{(int)} - The column size of the terminal to emulate (\ie{} size W).
    \item \texttt{dbfilename} - \texttt{(str)} - The path to the database file used. 
    \item \texttt{threadpool} - \texttt{(Object)} - An executor class that has a \texttt{map} function (\ie{} \texttt{concurrent.futures.ThreadPoolExecutor}. 
    \item{} \texttt{gameids} - \texttt{(List[str])} - A predetermined list of \textit{gameids} to iterate through. 
    \item{} \texttt{shuffle} - \texttt{(bool)} - Whether to shuffle the order of \textit{gameids} before iterating. 
    \item{} \texttt{loop\_forever} - \texttt{(bool)} - Whether to start again at the beginning of \textit{gameids} when the iterator runs out. If \texttt{False}, empty batch indexes are padded with \texttt{0}.
    \item{} \texttt{subselect\_sql} - \texttt{(str)} - A SQL string to dynamically generate subdataset, by select \textit{gameids}. 
    \item{} \texttt{subselect\_args} - \texttt{(Objects)} - Arguments to any SQL query provided in  \texttt{subselect\_sql}. 
\end{enumerate}

\subsection{Adding Datasets}
\label{app:addingdataset}

\begin{figure}
    \centering
        \caption{Example of how to simply create a database file and populate it with datasets. Note that each directory must be associated with a dataset name chosen by the user (\ie{} \texttt{``nld-aa''}). This is then used to access the data (as in Figure \ref{fig:threadpool}).}
    \includegraphics[width=0.8\linewidth]{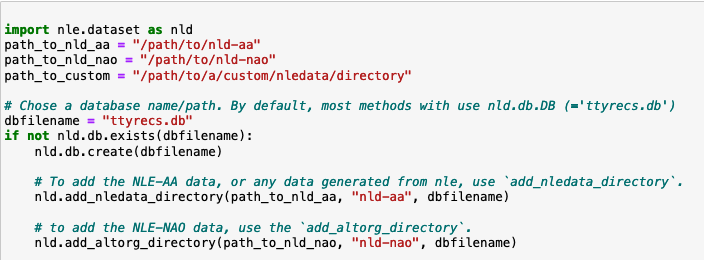}

    \label{fig:adding}
\end{figure}

Adding datasets to a the database is a very easy ``one-off'' operation. Examples are shown in the Tutorial Notebook and also in Figure \ref{fig:adding}.  

Since the process for adding \ttyrecthreebz{} and \ttyrecbz{} files are a little different from each other, two different methods should be used: \texttt{nle.dataset.add\_nledata\_directory} for \ttyrecthreebz{}, and \texttt{nle.dataset.add\_altorg\_drectory} for \ttyrecbz{} data from NAO. This is documented in the code and in the Tutorial Notebook.

\label{sec:appendix-populate}

\subsection{ThreadPool Usage}

\begin{figure}
    \centering
        \caption{Example of how to use a \texttt{ThreadPoolExecutor} using Python 3's built-in \texttt{concurrent.futures} module. This parallelism allows control of how many CPU workers are use to load data, without having to resort to Python's \texttt{multiprocessing}.}
    \includegraphics[width=0.8\linewidth]{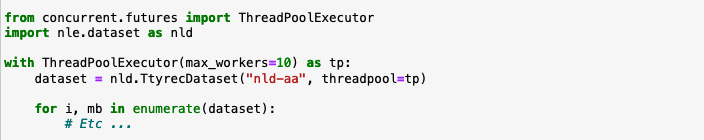}

    \label{fig:threadpool}
\end{figure}

All files are read independently, loaded into their own C++ \texttt{Converter} object that provides lightweight terminal emulation. This independence provides the potential for very effective parallelization along the batch index, B.  In the \texttt{Converter} object, we release the GIL, allowing for very true parallelization of file processing with Python's own threads. This can provide a many-fold boost to performance and throughput of frames per second, as can be seen in Table \ref{tab:threadpool}. Note, however, that the benefits of parallelization are mainly found with long sequence lengths. Example usage is shown in Figure \ref{fig:threadpool}.

\begin{table}[h]
\centering
\caption{The frames per second, mean (standard deviation), across 5 runs of 500k frames when served by the \NLD{} dataset with different parameters for batch size, sequence length, and number of CPUs used (which corresponds to the number of threads in the threadpool).}
\begin{tabular}{c | c c | c c c}
    \toprule
     & & & \multicolumn{3}{c}{Sequence Length} \\
     & & & 32 & 128 & 512 \\
    \midrule
     & \multirow{3}{6em}{Batch Size} & 32 & 20.6k (325) & 19.4k (250) & 20.7k (499) \\
    1 CPU & & 128 & 16.5k (159) & 18.9k (235) & 21.2k (254) \\
     & & 512 & 15.1k (325) & 17.1k (218) & 17.9k (94) \\
    \midrule
     & \multirow{3}{6em}{Batch Size}  & 32 & 73.8k (3.2k) & 110.8k (3.0k) & 142.7k (6.7k) \\
    10 CPUs & & 128 & 113.1k (1.5k) & 144.2k (1.9k) & 167.0k (3.3k) \\
     & & 512 & 114.2k (2.2k) & 129.8k (2.7k) & 139.1k (4.1k) \\
    \midrule
     & \multirow{3}{6em}{Batch Size}  & 32 & 86.2k (2.3k) & 134.6k (3.9k) & 231.2k (7.2k) \\
    80 CPUs & & 128 & 177.8k (2.9k) & 372.9k (16.5k) & 492.4k (24.4k) \\
     & & 512 & 222.4k (7.9k) & 446.7k (22.5k) & 482.5k (10.4k) \\
    \bottomrule
 \end{tabular}
\label{tab:threadpool}
 \vspace{-1.5em}
\end{table}

\subsection{SQL Usage}

\begin{figure}[h!]
    \centering
        \caption{Example of how to use a SQL queries to subselect a part of a dataset dynamically. This method was used to generate the \texttt{NLD-AA-Monk} runs only runs in Table \ref{tab:results} and Figure \ref{fig:training_curves}}
    \includegraphics[width=0.8\linewidth]{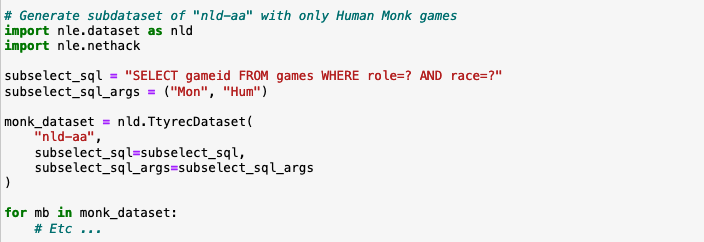}

    \label{fig:monksql}
\end{figure}

With such a large and diverse dataset, it is highly likely that many will wish to train on subselections of the data according to certain criteria, contained in the metadata.  The \DatasetObject{} enables this very easily, by exposing the ability to subselect \textit{gameids} on the basis of a SQL command.  An example of this is shown in Figure \ref{fig:monksql}, and was used to generate the plots in Figure \ref{fig:monksql}.

\subsection{Creating Custom Datasets}

\begin{figure}[h!]
    \centering
        \caption{Example of how to create a custom dataset using \NLE{} and \NLD{}. A variant of this method was used to create \NLDact{}.}
    \includegraphics[width=0.8\linewidth]{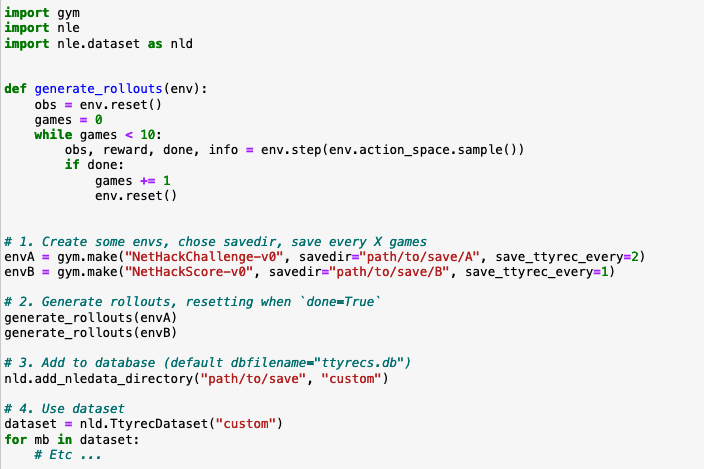}

    \label{fig:customdata}
\end{figure}

It is easy to create and add your own custom dataset using \NLE{} and \NLD{}. \NLE{} includes the ability to record files to a given save directory. The directory can then be directly added to NLD using the \texttt{add\_nledata\_directory} methods referred to in Appendix \ref{app:addingdataset}. An example is shown in Figure \ref{fig:customdata}.

\section{\NAO{} Details}
\label{sec:appendix-nao}

\begin{table}[]
    \centering
    \caption{A table of the conducts and achievements accomplished as a percentage of total episodes for each dataset. Most notably we can see only 1.46\% of episodes ascend in \NLDobs{}, and 0.04\% do this without any armour (Nudist). By contrast \NLDact{} achieves only the occasional solving of the Sokoban, indicating \NLDobs{} has a much better coverage of the late game.}
    \begin{tabular}{l r r}
     \toprule
   Conducts & \NLDact{} (\%) & \NLDobs{} (\%) \\
     \midrule
    Foodless & 0.89 & 21.16 \\
    Vegan & 1.35 & 29.50 \\
    Vegetarian & 1.41 & 32.18 \\
    Atheist & 2.36 & 52.13 \\
    Weaponless & 7.71 & 17.77 \\
    Pacifist & 0.26 & 8.53 \\
    Illiterate & 33.36 & 34.61 \\
    Polypileless & 100.00 & 96.88 \\
    Polyselfless & 87.65 & 93.59 \\
    Wishless & 98.91 & 95.88 \\
    Artifact wishless & 100.00 & 98.74 \\
    Genocideless & 100.00 & 96.90 \\
    \bottomrule
    \end{tabular}

    \begin{tabular}{l r r}
     \toprule
   Achievements & \NLDact{} (\%) & \NLDobs{} (\%) \\
     \midrule
    Got the Bell of Opening & 0.00 & 2.15 \\
    Entered Gehennom & 0.00 & 2.06 \\
    Got the Candelabrum of Invocation & 0.00 & 1.73 \\
    Got the Book of the Dead & 0.00 & 1.65 \\
    Performed the Invocation & 0.00 & 1.62 \\
    Got the Amulet of Yendor & 0.00 & 1.60 \\
    Was in the End Game & 0.00 & 1.56 \\
    Was on the Astral Plane & 0.00 & 1.53 \\
    Ascended & 0.00 & 1.46 \\
    Got the Luckstone at Mines' End & 0.00 & 3.95 \\
    Finished Sokoban & 0.17 & 5.93 \\
    Killed Medusa & 0.00 & 2.31 \\
    Zen conduct intact & 0.00 & 0.05 \\
    Nudist conduct intact & 0.00 & 0.04 \\
    \bottomrule
    \end{tabular}
    \label{tab:achievement}
\end{table}

\subsection{Conducts and Achievements}

By analysing the Conducts and Achievement flags in the \textit{conduct} and \textit{achieve} metadata fields, one can see the variety of behaviours in the \NLDobs{} dataset.
The aggregated \% of episodes with each conduct/achievement is shown for each dataset in Table \ref{tab:achievement}

\subsection{Metadata Matching Algorithm}
\label{app:metaassign}
On NAO, a game of NetHack can be saved and resumed, resulting in games split across several \ttyrecbz{} files and time periods. In these situations, only one entry is made to the communal \xlogfile{}, and this is made at the very end of the game with no link to which files are part of the game. 

However, it is possible to assign files to games making use of three pieces of information: first, that alt.org enforces that a user complete one game before starting another; second, that the \xlogfile{} records the \textit{starttime} and \textit{endtime} of the game; and, third, that each file is saved with the same filename template indicating the user and file creation time ( \texttt{<username>/<file\_creation\_timestamp>.ttyrec.bz2}).

Making use of these three pieces of information, we can assign the files to the games, by lining up all episodes per user in an \xlogfile{} by start and endtime, and then assigning files created between that start and endtime to that game. The full logic for this is carefully documented in the file \texttt{nle/datasets/populate\_db.py}.

In this way it is possible to remove games that have started but not finished from the alt.org S3 bucket, whilst also filter out games that may have started well before the first \textit{starttime} in an \xlogfile{} (indicating for whatever reason the game was not logged). All this logic is vital in ensuring that \NLD{} episodes can be seamlessly stitched together correctly from many file parts.

\subsection{Filtering `Bad' Episodes for \NLDobs{}}
\label{sec:filtering}
We filtered `bad' episodes for \NLDobs{} in three ways. First, we removed episodes from the dataset where the player seems to have participated in `start-scumming'. This practice involves starting a game of NetHack and ending it early if the randomly-generated starting attributes/inventory is not favourable enough. We filtered these games by filtering episodes from the dataset with fewer than 10 turns, where the death was `quit' or `escaped'.

We also filtered an episode with negative in-game \textit{turns} - something which should be impossible! This single player had managed to survive many orders of magnitude longer than any other game, eventually successfully overflowing the \textit{turns} counter (a 32-bit integer). While surviving this long is impressive, it is unlikely this episode demonstrate a drive to `ascend' and may pose a large bias to our dataset, and so is filtered out.

Finally we also decided to remove the files of players with highly offensive names from our dataset. These accounted for very few episodes and frames (less than 0.1\%) and largely from poor players. To add an extra layer of privacy, we pseudonymise all player names in the database, even though player names are generally already pseudonyms. 

\subsection{Noise Investigation}

The \NLDobs{} dataset is a real-world dataset collected over a decade, and as such contains some real world noise that can pose research problems.

The first problem is terminal rendering noise. Different terminals have different escape sequences and different characters sets, and some of these may not be interpretable by \NLD{}'s lightweight VT100 emulator.  While \NLD{} has made adjustments to support DEC graphics\footnote{\url{https://en.wikipedia.org/wiki/DEC_Special_Graphics}} and IBM graphics\footnote{\url{https://nethackwiki.com/wiki/IBMgraphics}}, it is still possible for a user to define their own exotic graphics which could render noisily in \NLD{}, perhaps with extra, unusual symbols. This problem is also faced by alt.org's own webplayer, but thankfully, this issues is only encountered very rarely.

Another source of rendering noise, aside from graphics, is terminal size. Players can play NetHack with different terminal sizes, and NetHack's clever paging system will adapt the screen to the right size. In our case we chose by default to render on the screen size used by \NLE{}, which is also the smallest "full size" you can play NetHack on (80 x 24). For episodes where the screen size was larger than this, we make the decision to crop the screen rather than wrap around the lines (as one typically would when emulating large commands on a small screen). It is possible to control the emulator size scren by using the \texttt{rows} and \texttt{cols} arguments to the constructor, if the user so wishes.

The second source of noise comes from errors introduced into the metadata assignment process, outlined in Appendix \ref{app:metaassign}. In some cases, such as under heavy traffic, NetHack can fail to write a line to an \xlogfile{}, and drop the line entirely. This introduces noise into the our assignment algorithm and could result in a user's games potentially including some \ttyrec{} files from a different one of their episodes.  We check for discrepancies between metadata and in-game play, by rendering the last 10 frames of all episodes in \NLDobs{} and checking whether the \textit{point} and number of \textit{turns} parsed from the screen match the gameplay metadata in \NLDobs{}. Since these two values are themselves not always displayed or might change after game end, this itself is a noisy process, yet we managed to obtain a upper bound on the errors. In the episodes we found at most \textasciitilde 5\% occurence of a discrepancy between the final frames of an episode and the metadata for \textit{score} and \textit{turns}, and were able to verify correct metadata for \textasciitilde 90\% of episodes (\textasciitilde 5\% we were unable to classify).

\section{Experiment Hyperparameters and Details}
\label{sec:appendix-experiments}

\subsection{Methods}
All methods are implemented using the open-source RL library \texttt{moolib}~\cite{moolib2022}. The online and online + offline RL methods are built on top of an APPO implementation based on~\cite{petrenko2020sample}. The offline RL methods are built on top of the DQN algorithm. For CQL, we followed the implementation from~\cite{d3rlpy}, which is an established library for offline RL algorithms. For IQL, we followed the original implementation open-sourced by the authors~\cite{iql}. 
We use the same base architecture and processing of the observations and actions for all algorithms. Please check out our open-sourced code for full details of these implementations. All experiments were run with 20 CPUs in approximately 1 day on single Volta 32GB GPUs.

\subsection{Preprocessing}
\label{app:preprocessing}
As a preprocessing step, a wrapper is used to render the observations \texttt{tty\_chars} and \texttt{tty\_colors} as pixels, which are then downsampled and cropped centered on \texttt{tty\_cursor}. The resulting image is added as an observation known as \texttt{screen\_image}.  This preprocessing step is standard for the Chaotic Dwarf baseline model that we use.~\cite{hambro2022insights} 

\subsection{Model}

We adapt the model from the standard Chaotic Dwarf Sample Factory baseline, that was open sourced\footnote{https://github.com/Miffyli/nle-sample-factory-baseline} during the NetHack Challenge.~\cite{hambro2022insights}  This model relies on three small encoders for observations, that combine their representations before passing through an LSTM and taking policy and baseline heads from the output. 

The original model relied on \NLE{}'s \texttt{blstats}, \texttt{message}, and the preprocessed \texttt{screen\_image} (outlined in Appendix \ref{app:preprocessing}) observations. Since the former two do not exist in the raw terminal output, instead they are replaced with their closest proxies - the top and bottom two lines of \texttt{tty\_chars} respectively. This involves slight changing to the normalization of the encoders. The full model can be found in our open sourced code.

\subsection{Hyperparameter Sweeps}
For DQN, we performed a hyperparameter search over $\epsilon_{start} \in [0.1, 0.2, 0.25, 0.5, 0.75, 0.8, 0.9]$, $\epsilon_{decay} \in [2500, 5000, 10000, 20000, 25000, 50000, 75000, 250000]$, $target$ $update$ $\in [40, 200, 400, 800, 2000, 4000]$ and tried both a MSE and a Huber loss. We found the following to work best: $\epsilon_{start}=0.25$, $\epsilon_{decay}=25000$, $target$ $update$ $= 400$, and MSE loss. For the common hyperparameters between CQL and IQL, we use the same values as the best ones found for DQN without additional tuning. For CQL, we also performed a hyperparameter search over $\tau \in [0.005, 0.05, 0.0005, 0.1, 0.01, 0.001, 0.0001]$ and for $cql$ $loss$ $coef$ $\in [1.0, 0.5, 0.1, 2.0, 10.0]$. We found $\tau = 0.005$ and $cql$ $loss$ $coef$ $= 2.0$ to work best. For IQL, we performed a hyperparameter search over $expectile \in [0.8, 0.7, 0.9]$ and $temperature \in [1.0, 0.1, 10.0, 0.5, 2.0]$. We found $expectile = 0.8$ and $temperature = 1.0$ to work best. Table~\ref{tab:hps} contains a list with all relevant hyperparameters used in our experiments (\ie{} the best ones found following our HP search). 
\begin{table}[t!]
    \centering
    \small
    \caption{List of hyperparameters used to obtain the results in this paper.}
    \begin{tabular}{cc}
    \toprule
    Hyperparameter & Value \\ [0.5ex]
    \midrule    
    activation function &  relu \\ [0.5ex]
    actor batch size &  512 \\ [0.5ex]
    adam betea1 &  0.9 \\ [0.5ex]
    adam beta2 &  0.999 \\ [0.5ex]
    adam eps &  0.0000001 \\ [0.5ex]
    adam learning rate &  0.0001 \\ [0.5ex]
    appo clip policy &  0.1 \\ [0.5ex]
    appo clip baseline &  1.0 \\ [0.5ex]
    baseline cost &  1  \\ [0.5ex]
    batch size &  256  \\ [0.5ex]
    crop dim &  18  \\ [0.5ex]
    grad norm clipping &  4 \\ [0.5ex]
    normalize advantages &  True \\ [0.5ex]
    normalize reward &  False \\ [0.5ex]
     num actor batches & 2 \\ [0.5ex]
     num actor cpus & 20 \\ [0.5ex]
     pixel size & 6 \\ [0.5ex]
     rms alpha & 0.99 \\ [0.5ex]
     rms epsilon & 0.000001 \\ [0.5ex]
     rms momentum & 0 \\ [0.5ex]
     reward clip & 10 \\ [0.5ex]
     reward scale & 1 \\ [0.5ex]
     unroll length & 32 \\ [0.5ex]
     use batchnorm & false \\ [0.5ex]
     use lstm & true \\ [0.5ex]
     virtual batch size & 128 \\ [0.5ex]
     rms reward norm  & true \\ [0.5ex]
     initialisation & orthogonal \\ [0.5ex]
     use global advantage norm & norm \\ [0.5ex]
     msg embedding dim & 32 \\ [0.5ex]
     msg hidden dim & 512 \\ [0.5ex]
     kickstarting loss & 1.0 \\ [0.5ex]
     ttyrec envpool size & 4 \\ [0.5ex]
     ttyrec batch size & 256 \\ [0.5ex]
     ttyrec unroll length & 32 \\ [0.5ex]
     ttyrec envpool size & 4 \\ [0.5ex]
     $\epsilon_{start}$ & 0.25 \\ [0.5ex]
     $\epsilon_{end}$ & 0.05 \\ [0.5ex]
     $\epsilon_{decay}$ & 25000 \\ [0.5ex]
     target update & 400 \\ [0.5ex]
     dqn loss & mse \\ [0.5ex]
     $\tau$ & 0.005 \\ [0.5ex]
     cql loss coef & 2.0 \\ [0.5ex]
     expectile & 0.8 \\ [0.5ex]
     temperature & 1.0 \\ [0.5ex]
    \bottomrule
    \end{tabular}
    \label{tab:hps}
\end{table}
\subsection{Evaluation}

We evaluate performance using the \texttt{eval.py} script in the experiment code. This script creates a \texttt{moolib.EnvPool} and assigns each batch index a number of episodes to run, to collect data. We collect 1024 episodes in this way, per run. 

It is important to note that this method of evaluating performance is slightly different to the one generating curves in Figure \ref{fig:training_curves}. For training curves, we add any episodes to a running mean as soon as they finish rolling out (\ie{} just before training). The end consequence of this is that short episodes with low scores are sampled disproportionately more often that very long episodes with high scores - in other words these curves are biased \textit{down}. Accurately sampling these very high scoring and very long rollouts account for the discrepancy between the curves in Figure \ref{fig:training_curves} and the values in Table \ref{tab:results}.

\section{On the Scale and Performance of \NLD{}}

We compare the scale and performance of \NLE{} and \NLD{} to  two other large scale datasets of human demonstrations: MineRL \cite{minerl} and StarCraft II \cite{starcraft2}.
In particular, we present the size of the datasets, in terms of trajectories, and benchmark the performance of the datasets and environments, measured in frames per seconds (FPS).  The overview can be found in Table \ref{tab:benchmark}. 

\begin{table}[h!]
\centering
\caption{The mean frames per second (± standard deviation) across 5 runs for each environment or dataset. Although batch sizes and sequence length can affect performance, it is clear that \NLD{} is considerably faster than rival datasets.}
\begin{tabular}{l c c c}
    \toprule
    Dataset     & \NLD{}     & StarCraft II & MineRL \\
    \midrule
    Trajectories      & \textbf{100,000 - 1,500,000} & 971,000 & <5000 \\
    \midrule
    Environment FPS & \textbf{26400} ± \textbf{1200} &  200-700            & 68 ± 6.2   \\
    Dataset FPS     & \textbf{113100} ± \textbf{1500  } &  1800 ± 2.9          &   5000 ± 89 \\
    \bottomrule
 \end{tabular}
\label{tab:benchmark}

\end{table}

\subsection{Overview}

These results support the claim of \NLD{} to be a large-scale dataset.  \NLDobs{} has more trajectories than dataset used in AlphaStar\cite{alphastar}, and possibly also more transitions, since the average episode of \NLE{} is longer than that of StarCraft II. Even \NLDact{} has an order of magnitude more trajectories than MineRL.

These results also support the claims of \NLD{} and \NLE{} to be very performant in terms of frames per second. The \NLE{} environment is at least an order of magnitude faster than either alternative environment, and \NLD{} is potentially several orders of magnitude faster at loading data. Further details in how the batch size and unroll length affect the performance of \NLD{} can be found in  Table \ref{tab:threadpool}.

\subsection{FPS Benchmarking Method}
   To compute the throughput of \NLD{} we used the threadpool with 80 threads on a machine with 80 CPUs, a batch size of 128 and a sequence length of 32, taking the mean and standard deviation from 5 runs to 500k frames. All assessments are done with \NLDact{}, the slower of the two datasets. For \NLE{} we ran the environment for 500k frames, with an agent sampling random actions and a max episode length of 1000 frames.  
   
   For StarCraft II we include the throughput of the environment as reported in the original paper, and measure the throughput of the dataset by running the \texttt{pysc2.bin.replay\_actions} function, provided in the github repo for batch processing replays, for 5 minutes with 80 threads on 80 cpus.  This too was repeated 5 times.
   
   For MineRL we used the \texttt{MineRLTreechop-v0} environment and dataset. We measured the environment speed by running headless on a machine with 80 cpus for 10k steps, sampling random actions and assuming a max episode length of 30k steps (note, the MineRL documentation suggests that running headless slows down the environment by a factor of 2-3x because the rendering is performed on the CPU instead of the GPU).  We measured the MineRL dataset throughput using the provided \texttt{BufferedBatchIter} with a batch size of 4096 for a full epoch. Once again, we repeated this 5 times.

   Table \ref{tab:benchmark} presents the mean and standard deviation of the frames per second.

\section{Limitations of our Work}
\label{app:limitations}
One potential limitation of our work is our human dataset \NLDobs{} doesn't contain any actions or rewards, so it cannot directly be used for imitation learning of offline RL. However, this is the case for many other domains of interests where you may have access to sequences of observations (\eg{} when learning skills from human or agent videos in domains like robotics or autonomous driving). Thus, we believe our human dataset can enable research on this more realistic and important setting in a safe environment. 

While \NLDact{} contains both actions and rewards thus allowing research on imitation learning and offline RL, it is less diverse than \NLDobs{} as it contains demonstrations from only one symbolic bot, the winner of the NetHack Challenge at NeurIPS 2022~\cite{hambro2022insights}. In the future, we hope the community will contribute many more datasets to \NLD{} collected using a variety of bots (including learning-based ones). We believe that the current dataset is a good starting point to develop even better agents at playing NetHack. Then, those agents could be used to collect additional trajectories which can be added to \NLD{}, resulting in a feedback loop that constantly grows the size and diversity of the dataset and develops better and better agents. 

Another potential limitation of our work is that we use the in-game score to train our RL agents, which is the standard practice from~\cite{nle} and~\cite{hambro2022insights}. While the in-game score roughly correlates with good performance on NetHack, it is not exactly the same as ascending \ie{} winning the game, which is ultimately how we want to evaluate agents. For example, humans sometimes attempt to ascend with as low a score as possible, as an additional challenge. This misalignment is common in other real-world domains where it can be difficult to design a good reward function that illicits the desired behavior (\eg{} autonomous driving). Thus, this is an open research challenge which could be enabled by our dataset, but lies outside the scope of the current work.

\section{Broader Impact Statement}
\label{app:broader-impact}
Our work introduces a new large-scale dataset based on a computer game to enable research in multiple domains including learning from demonstrations, imitation learning, and reinforcement learning. Our dataset is cheap to run in order to democratize research on large-scale datasets of demonstrations, which have historically been restricted to well-resourced (industry) labs.  This dataset contains human demonstrations, but these are already available online and the demonstrations are de-identified such that only a user's chosen usernames can be retrieved.
Thus, we don't envision any direct negative social impact of this work. Of course, people could use our dataset to develop new methods for learning from demonstrations. Since these techniques can be quite general, they could also be used for other real-world applications such as autonomous driving, robotics, healthcare, financial services, or recommendation systems (where large-scale datasets may be available). Of course, these domains are more sensitive than computer games, so much more care is needed to mitigate the risks and develop safe, robust systems. While we believe this lies outside the scope of current work, researchers and practitioners using our dataset should always keep an eye on potential negative societal impacts. 

\section{Corrections}

A prior version of this work used a model evaluation script that had a bug resulting in the inflation of all scores by a roughly constant scale, in Table \ref{tab:results}. This bug has been corrected and the results regenerated, using the same original checkpoints where possible.  For IQL, CQL and both DQN experiments, original checkpoints had been deleted, so experiments were rerun in full. Despite the unfortunate error, it should be noted that these results still support relevant findings of Section \ref{resultsanddiscussion} and the rest of the paper. Our thanks to Ulyana Piterbarg and Jens Tuyls for their help in spotting the bug.

\end{document}